\documentclass{article}
\usepackage[letterpaper,margin=1in]{geometry}
\usepackage{lmodern}
\usepackage[round]{natbib}

\usepackage[utf8]{inputenc} %
\usepackage[T1]{fontenc}    %
\usepackage{hyperref}       %
\usepackage{url}            %
\usepackage{booktabs}       %
\usepackage{amsfonts}       %
\usepackage{nicefrac}       %
\usepackage{microtype}      %
\usepackage{xcolor}         %
\usepackage{amsthm}
\usepackage{algorithm}
\usepackage{algpseudocode}
\usepackage{amsmath, amssymb, bm}
\usepackage[T1]{fontenc}
\usepackage[utf8]{inputenc}
\usepackage{tikz}
\usetikzlibrary{arrows.meta, decorations.pathreplacing}
\usepackage{graphicx}
\usepackage{subcaption}
\usepackage{float}

\usepackage{graphicx}
\usepackage{tikz}
\usepackage{amsmath,amssymb}
\usetikzlibrary{arrows.meta,positioning,fit,calc,matrix}

\definecolor{labone}{RGB}{214,35,129}
\definecolor{labtwo}{RGB}{35,92,196}
\definecolor{labthree}{RGB}{34,150,105}
\definecolor{softred}{RGB}{243,150,150}

\tikzset{
    >=Latex,
    state/.style={
        circle,
        draw,
        fill=white,
        minimum size=6.5mm,
        inner sep=0pt
    },
    prodbox/.style={
        draw,
        rounded corners=8pt,
        inner xsep=5pt,
        inner ysep=5pt
    },
    larrow/.style={
        ->,
        line width=0.9pt
    },
    labbox/.style={
        rounded corners=2pt,
        minimum width=7mm,
        minimum height=6mm,
        inner sep=1pt
    }
}

\usepackage{algorithm}
\usepackage{algpseudocode}
\makeatletter
\newenvironment{breakablealgorithm}
  {\begin{center}
     \refstepcounter{algorithm}
     \hrule height.8pt depth0pt \kern2pt
     \renewcommand{\caption}[2][\relax]{
       {\raggedright\textbf{\ALG@name~\thealgorithm} ##2\par}
       \ifx\relax##1\relax
         \addcontentsline{loa}{algorithm}{\protect\numberline{\thealgorithm}##2}
       \else
         \addcontentsline{loa}{algorithm}{\protect\numberline{\thealgorithm}##1}
       \fi
       \kern2pt\hrule\kern2pt
     }
  }
  {\kern2pt\hrule\relax
   \end{center}}
\makeatother

\def\arxivversion{}
\newcommand{\mainfigwidth}{\linewidth}

\begin{document}

\title{Monte Carlo Random Walk Kernel}
\author{Arslan Iamaletdinov\\HSE University \and Elfat Sabitov\\HSE University \and Konstantin Sorokin\\HSE University}
\date{}
\maketitle

\begin{abstract}
 We introduce the Monte Carlo Random Walk Kernel (MCRWK), an estimator whose per-pair cost is independent of graph size and derived directly from the random walk kernel (RWK) expansion. By normalizing the kernel coefficients, MCRWK samples a shared walk length and estimates its contribution using random walks on the input graphs. This gives a simple way to approximate RWK without constructing the product graph. For a fixed expected walk length and efficient neighbor sampling, each pairwise estimate takes \(O(m)\) time for \(m\) Monte Carlo samples, allowing the sample budget to control the trade-off between runtime and accuracy. We evaluate MCRWK against Graph Voyagers and classical RWK methods on synthetic and real graph datasets. In the reported experiments on large synthetic graphs, MCRWK is faster than Graph Voyagers and, at matched runtime, more accurate. It also has the lowest Gram-matrix runtime on the five real datasets studied. Code and experiments are available at \url{https://github.com/arslan-yam/random_walk_kernels_computations}
 
\end{abstract}

\section{Introduction}
\label{sec:intro}

Graphs are universal data structures that model relational information across diverse domains, including chemoinformatics (molecular graphs), bioinformatics (protein-protein interaction networks), neuroscience (connectomes), and social network analysis. A central task in these settings is to measure graph similarity for downstream problems including classification, clustering, and regression. Kernel methods address this task by defining a positive semi-definite kernel $K(G_1, G_2)$, which implicitly embeds graphs into a Hilbert space where inner products encode similarity.

In this paper, we consider \textbf{Random Walk Graph Kernels} (RWK). Given two unweighted graphs $\mathrm{G}_{1}$ and $\mathrm{G}_{2}$, with vertices $V(G_1) = \{v^1_1, ..., v^1_{N_1}\}$ and $V(G_2) = \{v^2_1, ..., v^2_{N_2}\}$ respectively, their product graph $\mathrm{G}_{1} \times \mathrm{G}_{2}$ is the direct product of vertices and edges of $\mathrm{G}_{1}$ and $\mathrm{G}_{2}$: 

\begin{align*}
V(G_1\times G_2)
  &= V(G_1)\times V(G_2),\\
E(G_1\times G_2)
  &= \bigl\{\bigl((u_1,u_2),(v_1,v_2)\bigr):\\
  &\qquad (u_1,v_1)\in E(G_1),\;
           (u_2,v_2)\in E(G_2)\bigr\}.
\end{align*}
Then, general RWK on $\mathrm{G}_{1} \times \mathrm{G}_{2}$ can be computed as:
\begin{equation}
\label{eq:rwk}
    k(G_1,G_2)
    =\mathbf{v}^{\top}\left[\sum_{k=0}^{\infty}
      \mu_{k}\mathbf{A}^{k}_{\times}\right]\mathbf{w}.
\end{equation}

Here $\mathbf{A}_{\times}$ is the normalized adjacency matrix of the product graph, $\{\mu_k\}_{k=0}^{\infty}$ is a non-negative coefficient sequence that defines kernel type, and $v,w\in \mathbb{R}^{N_1N_2}$ denote starting and stopping distributions over the product graph. These distributions are typically $\mathbf v= v_1\otimes v_2$ and $\mathbf w= w_1\otimes w_2$, where $\otimes$ is Kronecker product. 

The main computational bottleneck of RWKs is the product graph:
for two $N$-vertex graphs, it has $N^2$ vertices, and direct
computation costs $O(N^6)$.
\citet{VishwanathanEfficientKernel2010} reduced this to $O(N^3)$
using Sylvester equations.
More recently, \citet{ChoromanskiLinearKernels2025} introduced
Graph Voyagers (GVoys), the first linear-time algorithm for
general RWKs on sparse graphs.
GVoys avoid constructing the product graph by correlating
random walks on the two input graphs through shared random
variables, yielding unbiased estimates for labeled and
unlabeled graphs.

We propose Monte Carlo Random Walk Kernel (MCRWK), a simpler $O(m)$-time estimator for RWKs, where $m$ denotes the number of Monte Carlo samples. Like GVoys, MCRWK relies on synchronized random walks, but it samples the shared walk length directly from the normalized kernel coefficient distribution. This removes the need for auxiliary Rademacher variables and avoids anchor-point selection while preserving unbiasedness. 
The resulting algorithm is empirically faster on large graphs, while maintaining comparable approximation error to GVoys and unbiasedness. Our contributions are as follows: 
\textbf{(i)} We introduce MCRWK, a Monte Carlo estimator for random walk graph kernels with $O(m)$-time scaling;
\textbf{(ii)} its derivation for unlabeled and labeled graphs, including
geometric and exponential kernels;
\textbf{(iii)} proofs of unbiasedness and variance bounds showing that
the sample complexity does not depend on graph size; and
\textbf{(iv)} an empirical comparison with classical RWK baselines and
GVoys, demonstrating faster and more accurate computation on large graphs, complemented by evaluations
of Gram-matrix construction and downstream classification
and regression on real graph datasets.

\section{Related Work}
\paragraph{Exact Random Walk Kernel Computation.} The earliest approaches to RWKs constructed the direct product graph $G_1 \times G_2$ explicitly and compute $k_{\text{RWK}} = v^\top (I - \lambda A_\times)^{-1} w$ via matrix inversion \citep{gartner}. For graphs with $N$ vertices each, this yields $O(N^6)$ time and $O(N^4)$ memory complexity, and is therefore practical only for small graphs.

\paragraph{Cubic-Time Algorithms.} The first major algorithmic improvement came when \citet{Vishwanathan-early} proposed to compute RWK through Sylvester equations. The later work of \citet{VishwanathanEfficientKernel2010} proposed a unifying framework that exploits the Kronecker product structure of the product graph. When the product graph adjacency matrix factorizes as $\mathbf{A}_{\times} = \mathbf{P}_1 \otimes \mathbf{P}_2$, the kernel reduces to a Sylvester equation; for labeled graphs with $d$ edge labels, it reduces to a generalized Sylvester equation $A = \sum_{l=1}^{d} \lambda P_1^{(l)} A P_2^{(l)} + A_0$. The Sylvester approach runs in $O(N^3)$ time; conjugate gradient (CG) and fixed-point iteration (FPI) reduce this further on sparse graphs by exploiting fast matrix--vector products. The performance of these methods is ultimately bounded by the convergence of the underlying iterative solvers. Subsequent work refines the Sylvester \citep{fasten-sylvester-graphs,TIAN2024106674}, CG \citep{UpgradedCG} and FPI \citep{OseledetsFPI} subroutines, but none achieves linear time complexity.

\paragraph{Linear-time approximation: Graph Voyagers.}
The first step toward linear-time RWK computation was achieved by \citet{Kang2012FastRW} for the geometric kernel $\mu_k = \lambda^k$. More recently, \citet{ChoromanskiLinearKernels2025} introduced Graph Voyagers (GVoys), the first linear-time algorithm for general RWKs. GVoys avoids instantiating the product graph by sampling random walks on ${G}_1$ and ${G}_2$ and correlating them through shared Rademacher variables $g$ and $z$, which in expectation cancel out walks that mismatch in length or label. The resulting \textit{Graph Random Features} \citep{grf-choromanski23a} produce low-dimensional embeddings whose dot product is an unbiased estimate of the RWK. With $m$ walks and $r$ anchor points the time complexity is $\mathcal O\bigl(\frac{1}{p_{\text{halt}}} m \log(r) N + m r N\bigr)$, which is linear in $N$. Further works on GRFs introduce generalizations \citep{general-grf}, new optimized approaches to compute them \citep{grf++} and extend the applications of GRFs to new domains \citep{grf-swing-2026}.

\paragraph{Novel Monte Carlo approach.} 
Our proposed Monte Carlo Random Walk Kernel (MCRWK) shares the same goal of unbiased estimation, but adopts a different strategy. Instead of factorizing $\mu_k$ and introducing extra Rademacher variables to align walk lengths, we directly sample a shared walk length $K$ from the normalized distribution $\pi(k)=\mu_k/C$, where $C=\sum_{k=0}^{\infty}\mu_k$. This observation enables a plain Monte Carlo estimator $k = C\,\mathbb{E}[({v}_1^{\!\top}\mathbf{P}_1^{K}{w}_1)({v}_2^{\!\top}\mathbf{P}_2^{K}{w}_2)]$ that requires no deconvolution, no $g$ or $z$ variables, and no anchor point selection. For labeled graphs, an extra importance-sampling step over label sequences replaces the complicated $z$-variable mechanism, allowing flexible tuning of the proposal distribution $q$.

\section{Methods}
\label{sec:methods}
We now derive MCRWK for unlabeled and labeled graphs. Let $G_1$ and $G_2$ be two graphs with adjacency matrices $A_1$ and $A_2$. Their normalized adjacency matrices are $P_1 = D_1^{-1} A_1$, $P_2 = D_2^{-1} A_2$, where $D_i$ is a diagonal matrix with vertex degrees on the diagonal. As in Equation~\ref{eq:rwk}, the RWK is computed on the product graph $G_1\times G_2$. In the unlabeled case, its transition matrix is $A_\times=P_1\otimes P_2$. Write $v=v_1\otimes v_2$ and $w=w_1\otimes w_2$.

Let $\{\mu_k\}_{k=0}^{\infty}$ be a non-negative sequence with finite sum 
$C := \sum_{k=0}^{\infty} \mu_k < \infty$. We define a random variable $K$ 
via the probability function
$\mathbb{P}(K = k) = \frac{\mu_k}{C}, k \in \mathbb{N} \cup \{0\}$. The random walk kernel takes the form
\begin{align}
k(G_1,G_2)
  = \sum_{k=0}^{\infty}\mu_k\,v^\top A_\times^k w \notag
  = C\,\mathbb E_K[v^\top A_\times^K w].
\label{eq:mcrwk}
\end{align}

Two natural choices of $\{\mu_k\}$ yield standard kernels:

\textbf{Geometric kernel.} Setting $\mu_k = \lambda^k$ for $\lambda \in (0,1)$ gives 
$C = \frac{1}{1-\lambda}$ and $K \sim \mathrm{Geom}(1-\lambda)$:
\[
k(G_1,G_2)=\frac{1}{1-\lambda}\,
  \mathbb E_K[v^\top A_\times^K w].
\]

\textbf{Exponential kernel.} Setting $\mu_k = \frac{\lambda^k}{k!}$ gives $C = e^{\lambda}$ 
and $K \sim \mathrm{Pois}(\lambda)$:
\[
k(G_1,G_2)=e^\lambda\,\mathbb E_K[v^\top A_\times^K w].
\]

\subsection{Unlabeled case.}\label{seq:unlabeled_case}

When the product graph adjacency matrix factorizes as $A_{\times} = P_1 \otimes P_2$, the mixed-product 
property of the Kronecker product gives $(P_1 \otimes P_2)^k = P_1^k \otimes P_2^k$, 
and the kernel can be written as
\begin{align}
k(G_1,G_2)
  &= C\,\mathbb E_K\!\left[v^\top
      (P_1^K\otimes P_2^K)w\right] \notag\\
  &= C\,\mathbb E_K\!\left[
      (v_1^\top P_1^K w_1)
      (v_2^\top P_2^K w_2)\right].
\label{eq:mcrwk-unlabeled}
\end{align}
This factorization allows us to simulate 
independent random walks on $G_1$ and $G_2$ separately, without explicitly constructing the product graph. By sharing the random walk length $K$ across graphs, the estimator remains unbiased while achieving complexity that is independent of graph size. Moreover, the walks for each graph are simulated once and reused across all pairs, so the random walk simulation scales linearly with the number of graphs in the dataset.

\subsection{Labeled case}\label{seq:labeled_case}

The unlabeled construction extends naturally to labeled graphs. When edges carry 
labels from a set $[d] := \{1, \dots, d\}$, the weight matrix takes the form $ A_{\times} = \sum_{l=1}^d P_1^{(l)} \otimes P_2^{(l)},$
where $P_s^{(l)}$ is the normalized adjacency matrix of graph $G_s$ restricted to edges 
with label $l$. For a label sequence $i=(i_1,\ldots,i_K)$, write
$P_s^i=P_s^{(i_1)}\cdots P_s^{(i_K)}$. Expanding $A_\times^K$ gives
\begin{align}
k(G_1,G_2) = 
   C\,\mathbb E_K\Bigl[
      \sum_{i\in[d]^K}(v_1^\top P_1^i w_1)
      \notag \cdot(v_2^\top P_2^i w_2)\Bigr].
\label{eq:mcrwk-labeled-expansion}
\end{align}

For a fixed label sequence ${i} = (i_1, \dots, i_K) \in [d]^K$, the 
term $v_s^\top P_s^{(i_1)} \cdots P_s^{(i_K)} w_s$ is simply the expected 
terminal weight of a random walk on $G_s$ where step $t$ is restricted to 
edges with label $i_t$. Writing $\phi_s({i})$ for this quantity, the 
kernel becomes
\[
    k(G_1, G_2) = C\,\mathbb{E}_K\!\left[\sum_{{i} \in [d]^K} 
    \phi_1({i})\,\phi_2({i})\right]
\]

Summing over all $d^K$ label sequences is computationally expensive, so we approximate this sum via importance sampling. Let $q = (q_1, \dots, q_d)$, $q_l > 0$, 
$\sum_l q_l = 1$, be a distribution over labels, and sample 
$L_1, \dots, L_K \sim q$. The probability of a 
sequence ${i}$ under $q$ is $q({i}) = \prod_{t=1}^K q_{i_t}$, 
so the sum rewrites as an expectation:
\begin{align*}
\sum_{i\in[d]^K}\phi_1(i)\phi_2(i)
  &=\mathbb E_L\!\left[
      \frac{\phi_1(L)\phi_2(L)}{q(L)}\right]\\
  &=\mathbb E_L[F_1(L)F_2(L)],
\end{align*}
where we define $F_s({L}) := \frac{\phi_s({L})}{\sqrt{q({L})}}$
to symmetrically split the importance weight between the two graphs.
The kernel then takes the form

\begin{equation}
\label{eq:mcrwk-lableled}
    k(G_1, G_2) = C\,\mathbb{E}_K\,\mathbb{E}_{{L}}\!\left[
    F_1({L})\, F_2({L})\right],
\end{equation}
which we estimate via Monte Carlo by sampling $K \sim \frac{\mu}{C}$ and 
$L_1, \dots, L_K \sim q$.

\begin{algorithm}
\caption{Walk function}
\begin{algorithmic}
\Function{Walk}{$P_s, v_s, w_s, L = (L_1,\dots,L_K)$}
    \State Sample $x_0 \sim v_s$, \quad $\texttt{weight} \gets 1$
    \For{$t = 1, \dots, K$}
        \State $\texttt{weight} \gets \texttt{weight} \times \displaystyle\sum_{y} P_s^{(L_t)}[x_{t-1}, y]$
        \If{$\texttt{weight} = 0$}
            \State \Return $0$
        \EndIf
        \State $x_t \sim \operatorname{normalize}\big(P_s^{(L_t)}[x_{t-1}, \cdot\,]\big)$
    \EndFor
    \State \Return $\texttt{weight} \times w_s(x_K)$
\EndFunction
\end{algorithmic}
\end{algorithm}

\begin{algorithm}[!t]
\caption{Monte Carlo estimation of Gram matrix $K \in \mathbb{R}^{N_G \times N_G}$ --- labeled case}\label{alg:mcrwk}
\begin{algorithmic}[1]
\Require For each $G_i$, label-indexed transition matrices
$\{P_i^{(l)}\}_{l=1}^d$, initial distribution $v_i$ and stopping
distribution $w_i$; importance sampling distribution
$q=(q_1,\dots,q_d)$ over
$\mathcal L=\bigcup_{i=1}^{N_G}\mathrm{labels}(G_i)$;
coefficient sequence $\mu_k$ with sum $C$;
$m$ length samples and $n$ label samples per length
\Ensure Gram matrix $\hat K\in\mathbb R^{N_G\times N_G}$
with $\hat K_{ij}\approx k(G_i,G_j)$
\State Sample shared walk lengths $K_1, \dots, K_m \sim \mathbb{P}(K = k) = \dfrac{\mu_k}{C}$
\For{$i = 1, \dots, m$}
    \For{$j = 1, \dots, n$}
        \State Sample label sequence $L^{(i,j)} = (L_1, \dots, L_{K_i}) \overset{\text{i.i.d.}}{\sim} q$
        \State Compute $q(L^{(i,j)}) = \displaystyle\prod_{t=1}^{K_i} q_{L_t^{(i,j)}}$
    \EndFor
\EndFor
\For{$i = 1, \dots, N_G$}
    \For{$p = 1, \dots, m$}
        \For{$j = 1, \dots, n$}
            \State $\hat{\phi}_i(L^{(p,j)}) \gets \textsc{Walk}(P_i, v_i, w_i, L^{(p,j)})$
            \State $F_i^{(p,j)} \gets \dfrac{\hat{\phi}_i(L^{(p,j)})}{\sqrt{q(L^{(p,j)})}}$
        \EndFor
    \EndFor
\EndFor
\For{$i = 1, \dots, N_G$}
    \For{$j = 1, \dots, i$}
        \State $\hat{K}_{ij} \gets \hat{K}_{ji} \gets \dfrac{C}{mn}
        \displaystyle\sum_{p=1}^{m}\sum_{s=1}^{n} F_i^{(p,s)} \cdot F_j^{(p,s)}$
    \EndFor
\EndFor
\State \Return $\hat{K}$
\end{algorithmic}
\end{algorithm}

\paragraph{Unbiasedness.}
By construction, the MCRWK estimator is unbiased for any non-negative coefficient sequence $\{\mu_k\}$ with finite sum $C$, provided that the two factors are sampled independently:
\[
\mathbb{E}\bigl[\hat k(G_1,G_2)\bigr] = k(G_1,G_2).
\]
For diagonal entries, Algorithm~\ref{alg:mcrwk} reuses the same sample for both factors, introducing an upward bias:
\begin{align}
    \mathbb{E}\bigl[\hat k(G_1,G_1)\bigr]
&= \frac{C}{mn}\,\mathbb{E}\bigl[\lVert F(G_1)\rVert^2\bigr] \notag\\
&= k(G_1,G_1)
  + C\,\mathbb{E}\bigl[\operatorname{Var}(F\mid K,L)\bigr],
\end{align}
where the conditional variance is taken over the walk, at a fixed walk length
$K$ and label sequence $L$ (in the unlabeled case the conditioning on $L$ is
vacuous).
Keeping this diagonal yields a positive semi-definite Gram matrix. Alternatively, we obtain an unbiased diagonal by generating
two independent copies, $F_1(G_1)$ and $F_2(G_1)$, of $F(G_1)$
and using their product instead of the squared sample.
The resulting estimator is unbiased, since
\begin{equation}
    \label{eq:diag}
    \mathbb{E}\left[
        \frac{C}{mn}F_1(G_1)F_2(G_1)
    \right]
    = k(G_1,G_1).
\end{equation}

\subsection{Variance Bounds}
\label{sec:variance_bounds}
\paragraph{Unlabeled case}
For terminating distribution $w$ let $w_\infty := \max_i \|w_i\|_\infty \leq 1$. Each feature
product $F_{1,s} F_{2,s}$ (Alg.~\ref{alg:mcrwk}) is non-negative and bounded by $w_\infty^2$,
since $F_{i,s} = w_i(x^{(i,s)}_{K_s}) \in [0, w_\infty]$. As the $m$
samples are i.i.d.,
\begin{align}
\mathrm{Var}\bigl(\hat k(G_1,G_2)\bigr)
  &=\frac{C^2}{m}\,
    \mathrm{Var}(F_{1,1}F_{2,1})\notag\\
  &\leq\frac{C^2w_\infty^4}{m}
   \leq\frac{C^2}{m}.
\label{var-unlabeled}
\end{align}
Since the estimator is unbiased, this is also a bound on its mean
squared error. The standard deviation decays at a rate
$\sigma(\hat k) = O\bigl(C / \sqrt{m}\bigr)$

Applying Hoeffding's inequality gives
\begin{equation}
\label{Hoeffding-unlabeled}
\mathbb{P}\bigl(|\hat k - k| > \epsilon\bigr)
\;\leq\;
2 \exp\!\left(-\frac{2 m \epsilon^2}{C^2\, w_\infty^4}\right).
\end{equation}

So to achieve absolute error at most $\epsilon$ with probability at least
$1 - \delta$ it suffices to take
\begin{equation}
\label{sampling-unlabled}
m \;\geq\; \frac{C^2\, w_\infty^4}{2\,\epsilon^2}\,\log\frac{2}{\delta}.
\end{equation}
This bound is {independent of the graph size $N$, the
edge count $|E|$, and the dataset size $N_G$}. The sample budget
needed for a target precision depends only on the kernel parameter, the magnitudes of the boundary distributions, and the target precision.

\paragraph{Labeled case}
Let $q_{\min}=\min_l q_l$ and
$R(q)=\mathbb E_K[q_{\min}^{-K}]$.
For $m$ walk lengths and $n$ label sequences per length,
\begin{equation}
\mathrm{Var}\bigl(\hat k(G_1,G_2)\bigr)
\leq \frac{C^2w_\infty^4}{m}
\left(\frac{R(q)}{n}+\frac14\right).
\label{var-labeled}
\end{equation}
Chebyshev's inequality then gives
\begin{equation}
\mathbb P\bigl(|\hat k-k|>\epsilon\bigr)
\leq \frac{C^2w_\infty^4}{m\epsilon^2}
\left(\frac{R(q)}{n}+\frac14\right),
\label{Chebyshev-labeled}
\end{equation}
so absolute error at most $\epsilon$
with probability at least $1-\delta$ is guaranteed whenever
\begin{equation}
m\geq \frac{C^2w_\infty^4}{\delta\epsilon^2}
\left(\frac{R(q)}{n}+\frac14\right).
\label{labeled_sample_complexity}
\end{equation}
As in the unlabeled case, the required budget has no explicit
graph-size dependence.

Here $n$ controls the label-sampling variance, while $m$
also controls the walk-length variance, which cannot be
reduced by increasing $n$ alone.
For a fixed walk budget $B=mn$, $n=1$ is optimal.
The derivation and the budget comparison are given in
Appendix~\ref{app:labeled-variance}.

For the geometric kernel,
$R(q)=(1-\lambda)/(1-\lambda/q_{\min})$ is finite when
$\lambda<q_{\min}$.
For the exponential kernel,
$R(q)=\exp\!\bigl(\lambda(q_{\min}^{-1}-1)\bigr)$
is always finite.
Uniform $q_l=1/d$ minimizes this worst-case bound,
although frequency-based proposals can reduce the
actual variance in practice.

\subsection{Pairwise complexity}
\label{sec:pairwise_complexity}

\paragraph{Unlabeled case.}
Let $\bar K := \mathbb{E}[K]$ be the
expected walk length under the sampling distribution. For the two standard kernels 
$\bar K = \tfrac{\lambda}{1-\lambda}, \bar K = \lambda$
both independent of the graph size $N$, so $\bar K = O(1)$ for fixed
kernel parameter $\lambda$. Consider two graphs $G_1, G_2$ with $N_1, N_2$ vertices and average
degrees $\bar d_1, \bar d_2$. Sampling the $m$ shared walk lengths
$K_1,\dots,K_m$ in line~1 of Algorithm~\ref{alg:mcrwk_unlabeled} (Appendix~\ref{ap:algorithm}) costs $O(m)$. On graph $G_s$,
simulating a walk of length $K_i$ requires $K_i$ neighbour-sampling
steps, each performed in $O(\bar d_s)$ time when $P_s$ is stored in sparse
(CSR) format, a single row of $P_s$ has at most $\bar d_s$ non-zero
entries on average. Summing over all $m$ walks and taking the
expectation over $K$,

\begin{equation}
    T_{\text{walks}}(G_s) \;=\; O\bigl(m\,\bar K\, \bar d_s\bigr)
\end{equation}

The estimate $\hat k(G_1,G_2) = \tfrac{C}{m} \langle F_1,F_2 \rangle$ is one
dot product costing $O(m)$. The total pairwise time is

\begin{equation}
    T_{\text{pair}}^{\text{unl}} \;=\; O\bigl(m\,\bar K\,(\bar d_1+\bar d_2)\bigr).
\end{equation}

For sparse graphs of bounded degrees we have $\bar d_1, \bar d_2=O(1)$, therefore
$T_{\text{pair}}^{\text{unl}} = O(m\bar K)$ is {independent of the
number of vertices~$N$}. If one lets the sample budget scale with graph size, $m = cN$ for some constant $c>0$ the per-pair cost becomes $T_{\text{pair}}^{\text{unl}}
= O(cN\,\bar K) = O(N)$, recovering the linear scaling of
GVoys~\citep{ChoromanskiLinearKernels2025} but without the $1/p_{\text{halt}}$,
$\log r$, or anchor-point factors.

\paragraph{Labeled case.}
Algorithm~\ref{alg:mcrwk} additionally samples $n$ label sequences per walk length, so the total pairwise cost is 

\begin{equation}
    T_{\text{pair}}^{\text{lab}} \;=\;
O\bigl(m\,n\,\bar K\,(\bar d^{(\ell)}_1+\bar d^{(\ell)}_2)\bigr).
\end{equation}

This can be free of any explicit
$N$-dependence on sparse graphs, or linear if we let $m = cN$. Also in contrast to the $O(d^K)$ cost of summing over label sequences exactly this is linear in $\bar K$

\paragraph{Memory.}
MCRWK does not build the product graph. It stores the sparse
adjacency matrices, $O(|E_1|+|E_2|)$, and the feature vectors $F_i \in
\mathbb{R}^m$, so

\begin{equation}
    S_{\text{pair}} \;=\; O\bigl(|E_1|+|E_2|+m\bigr)
\end{equation}

versus $O(N^4)$ for the direct method and $O(N^2)$ for Sylvester /
CG / FPI, and $O\bigl(|E_1|+|E_2|+mr\bigr)$ for GVoys, whose feature
vectors carry an extra factor of $r$ from the anchor points.

\begin{table}[!t]
\centering
\caption{Pairwise RWK time and space complexity.
$N$: vertices per graph; $|E|$: edges; $\bar d$: average degree; $d$: number of edge labels;
$m$: Monte Carlo samples; $\bar K=\mathbb E[K]$: expected
walk length. On sparse graphs of bounded degree, MCRWK time
is independent of $N$.}
\label{tab:complexity}
\ifdefined\arxivversion
\begin{tabular*}{\textwidth}{@{}l@{\extracolsep{\fill}}ll@{}}
\toprule
Method & Pairwise time & Space \\
\midrule
Direct & $O(N^6)$ & $O(N^4)$ \\
Sylvester & $O(N^3)$ & $O(N^2)$ \\
CG / FPI & $O(k\,|E|)$ & $O(N^2)$ \\
GVoys & $O\bigl(mrN + \tfrac{1}{p_{\text{halt}}}\,m\log(r)\,N\bigr)$ & $O(|E|+mr)$ \\
\textbf{MCRWK} & $O(m\,\bar K\,\bar d)$ & $O(|E|+m)$ \\
\bottomrule
\end{tabular*}
\else
\small
\setlength{\tabcolsep}{3pt}
\begin{tabular}{@{}lll@{}}
\toprule
Method & Pairwise time & Space \\
\midrule
Direct
  & $O(N^6)$
  & $O(N^4)$ \\
Sylvester
  & $O(N^3)$
  & $O(N^2)$ \\
CG / FPI
  & $O(k\,|E|)$
  & $O(N^2)$ \\
GVoys
  & $\begin{aligned}[t]
       &O\!\bigl( mrN +\\
       & \dfrac{1}{p_{\text{halt}}} \,m\log(r)\,N\bigr)
     \end{aligned}$
  & $O(|E|+mr)$ \\
\textbf{MCRWK}
  & $O(m\,\bar K\,\bar d)$
  & $O(|E|+m)$ \\
\bottomrule
\end{tabular}
\fi
\end{table}

\subsection{Dataset complexity}
\label{sec:dataset_complexity}

A central practical advantage of MCRWK on a dataset
$\mathcal D = \{G_1,\dots,G_{N_G}\}$ is that the walk lengths -- and,
in the labeled case, the label sequences -- are sampled \emph{once}
and reused across all graphs. Once per-graph Monte Carlo features are
precomputed, every kernel evaluation reduces to a dot product,
independent of graph size.

For each graph $G_i$ the feature vector $F_i \in \mathbb{R}^{mn}$ takes
$T_{\text{feat}} = O(mn\,\bar K\, \bar d)$ to compute, with $\bar d$ the
average degree (in the unlabeled case $n = 1$). So for the
dataset the time complexity for computing features is
$O(N_G\, mn\, \bar K\, \bar d)$. This step is linear in $N_G$ and is
parallelizable across graphs.

Given the precomputed features, the kernel value for any pair is
$\hat K_{ij} = \tfrac{C}{mn}\langle F_i, F_j\rangle$, which takes
$T_{\text{pair}} = O(mn)$ time. This makes MCRWK particularly suited to
settings where the same dataset is queried repeatedly. Storing the input
graphs and precomputed features requires $O\bigl(N_G\, mn + \sum_i |E_i|\bigr)$.

Stacking the features into a matrix
$\mathbf F = [F_1, \dots, F_{N_G}]^\top \in \mathbb{R}^{N_G \times mn}$,
the off-diagonal part of the approximate Gram matrix is given by
\begin{equation}
\hat{\mathbf K} = \frac{C}{mn}\,\mathbf F \mathbf F^\top,
\label{eq:gram-lowrank}
\end{equation}
which has rank at most $mn$. So MCRWK gives a low-rank factorization of
the Gram matrix in time linear in $N_G$. This avoids rerunning a pairwise
RWK solver for each of the $O(N_G^2)$ graph pairs. The explicit Gram
matrix costs $O(N_G^2\, mn)$, but it is often not needed, since kernel methods
such as ridge regression, SVM or kernel PCA can work directly with
$\mathbf F$ as a linear model on explicit features, in time linear in $N_G$.

There are two options for the diagonal. First, the diagonal entries can
be replaced by the unbiased estimate~\eqref{eq:diag}, computed from an independent copy of the features. Then the Gram matrix is the sum
of a low-rank and a diagonal matrix, but it is no longer guaranteed to be
positive semi-definite. Second, the diagonal of~\eqref{eq:gram-lowrank}
can be kept as is. Its bias is
$\tau_i = C\,\mathbb{E}[\mathrm{Var}(F_i \mid K, L)] \geq 0$, so
$\mathbb{E}[\hat{\mathbf K}] = \mathbf K + \mathrm{diag}(\tau_1,\dots,\tau_{N_G})$.
This works as a graph-dependent ridge regularization and keeps
$\hat{\mathbf K}$ positive semi-definite.

\providecommand{\mainfigwidth}{0.8\linewidth}
\section{Experiments}\label{sec:experiments}
We compare MCRWK against five baselines: direct, conjugate
gradient (CG), fixed-point iteration (FPI), the Sylvester-equation method and
GVoys.

\paragraph{Setup.}
Unless stated otherwise we use the geometric kernel $\mu_k=\lambda^k$ with $\lambda=0.7$, so $C=1/(1-\lambda)\approx 3.3$ and the
expected walk length is $\bar K=\lambda/(1-\lambda)\approx 2.3$. To avoid the
degenerate kernel (Appendix~\ref{app:degenerate}) we draw
$v_i=|z_i|/\sum_j|z_j|$ with $z_j\sim\mathcal N(0,1)$ and $w$ independently in
the same way; all methods use the same $v$ and $w$. In the labeled case MCRWK samples label sequences from the uniform proposal $q$,
which maximizes $q_{\min}$ and therefore gives the bound of
Section~\ref{sec:variance_bounds} the best chance of being finite, with $n=1$ sequence per length,
which is optimal for a fixed budget. Appendix~\ref{ap:q_sampling} shows that
frequency-based proposals are empirically more accurate. MCRWK is run in two regimes: a
fixed budget $m\in\{10^2,10^3,10^4\}$, independent of the graph size as the
size-free bounds suggest, and a linear budget $m=cN$, whose constant $c$ is
calibrated so that MCRWK and GVoys take the same time. The baselines, the symmetric
transform through which GVoys is applied to our non-symmetric $P=D^{-1}A$, the
protocols and the hardware are described in Appendix~\ref{app:setup}.

\subsection{Scaling with graph size}
\label{sec:scaling-with-graph-size}

\begin{figure*}[!t]
  \centering
  \includegraphics[width=\mainfigwidth]{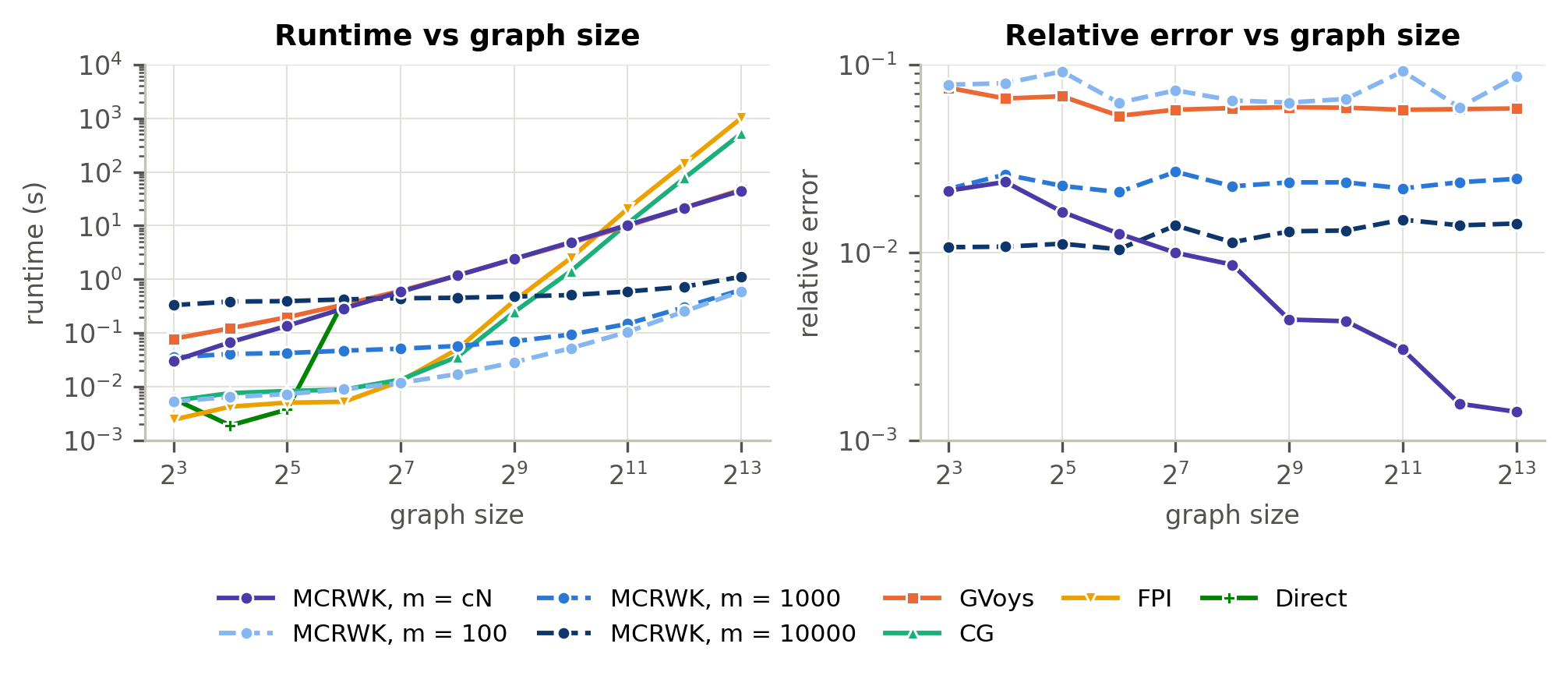}
  \caption{Runtime (left) and relative error $|\hat k-k|/k$ (right) of a single
  kernel value $k(G_1,G_2)$ vs graph size, for pairs of labeled random graphs with $d=3$ edge labels, averaged over four graph families. MCRWK with $m=cN$ and GVoys
  have matching runtimes by construction, so their curves overlap on the left.}
  \label{fig:scaling_labeled}
\end{figure*}

We first measure the cost and accuracy of a single kernel evaluation as the
graphs grow. For every size $N\in\{2^3,\dots,2^{13}\}$ we draw pairs of graphs from four random-graph families (Barabasi--Albert, Erdos--Renyi, Watts--Strogatz and a stochastic block model; Appendix~\ref{app:setup}) and average over them. In
the labeled setting every edge receives one of $d=3$ labels uniformly at random.
Each method computes only the pair value $k(G_1,G_2)$, so the reported time is
the per-pair cost. Figure~\ref{fig:scaling_labeled} shows runtime and relative
error.

\paragraph{Runtime.}
At $N=8192$ MCRWK with the three fixed budgets takes 0.60, 0.64 and 1.12\,s,
while GVoys takes 46.8\,s, CG 519\,s and FPI about $1000$\,s. MCRWK
with $m=cN$ takes 44.8\,s, matching GVoys by construction. MCRWK with $m=10^4$
is thus about $460\times$ faster than CG and $900\times$ faster than FPI.

\paragraph{Accuracy.}
With a fixed budget the error of MCRWK does not depend on the graph size: it
stays within 5.9--9.2\%, 2.1--2.7\% and 1.0--1.5\% for $m=10^2,10^3,10^4$ across
three orders of magnitude of $N$, matching the size-free sample complexity, and it falls by a factor of $3.2$ from $m=10^2$ to $10^3$, as $\sigma(\hat k)=O(C/\sqrt m)$ predicts, though only by $1.9$ at the next tenfold step, where the heavy-tailed labeled weights slow the rate (Appendix~\ref{ap:convergence}). At equal runtime MCRWK is more accurate
than GVoys, which keeps an error of 5.3--7.5\% at every size, 2.8 to 41 times that of MCRWK with $m=cN$. Unlabeled graphs behave the same way
(Appendix~\ref{ap:synthetic}).

\subsection{Gram matrix computation}
\label{sec:exp-gram}

\begin{figure*}[!t]
  \centering
  \includegraphics[width=\mainfigwidth]{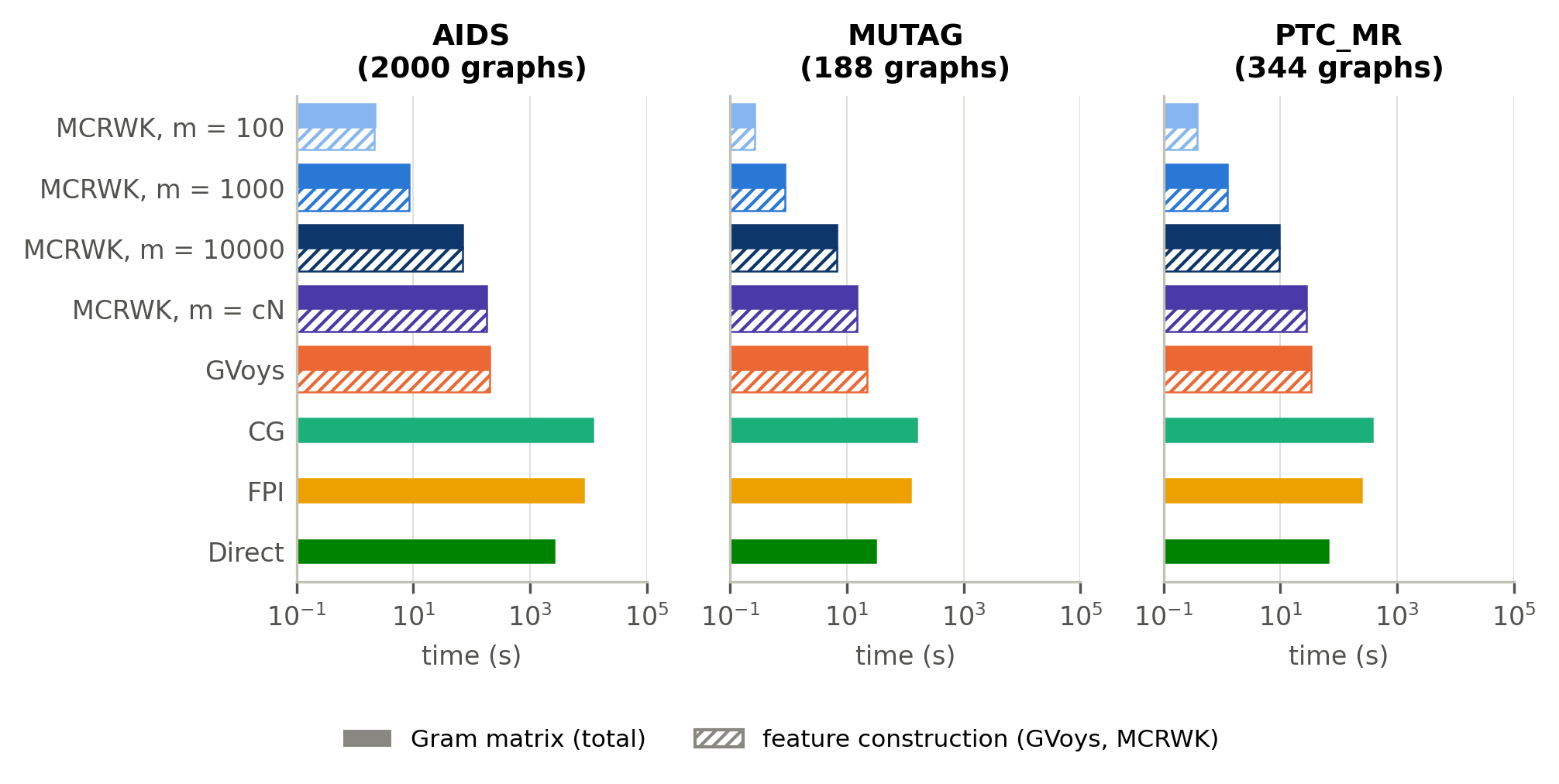}
  \caption{Time to compute the Gram matrix of a whole dataset with edge labels}
  \label{fig:gram_time_labeled}
\end{figure*}

Kernel methods require a full dataset Gram matrix rather than a single kernel
value. We measure its computation time on three edge-labeled TU
datasets~\citep{tudataset}: AIDS (2000 graphs), MUTAG (188) and PTC\_MR (344).
Exact methods solve $O(N_G^2)$ linear systems, one per graph pair, while GVoys
and MCRWK compute features once per graph and form the Gram matrix by one matrix
multiplication (Section~\ref{sec:dataset_complexity});
Figure~\ref{fig:gram_time_labeled} reports feature construction and total time
separately. The unlabeled case, including ENZYMES and NCI1, is in
Appendix~\ref{app:gram_time}.

Fixed-budget MCRWK is fastest on all three datasets. On AIDS its Gram times are
2.3, 9.0 and 73\,s for $m=10^2,10^3,10^4$, against $2.5\times10^3$\,s for
Direct, the fastest exact method, $8.4\times10^3$\,s for FPI and
$1.2\times10^4$\,s for CG: $m=10^4$ is $35\times$ faster than Direct and
$160\times$ faster than CG, and $m=10^2$ over $1000\times$ faster than Direct.
The gap is smaller but still substantial on MUTAG and PTC\_MR, where MCRWK takes
0.27--6.8\,s and 0.38--9.5\,s across the three budgets, against 31\,s and 66\,s
for Direct. Feature construction accounts for over 98\% of the total time of
both MCRWK and GVoys.

\subsection{Application: learning on labeled graphs}
\label{sec:exp-app}

Finally we use the kernels for classification on MUTAG (188 graphs) and regression of the constrained
solubility on ZINC\_test (5000 graphs), both with bond types as edge labels. We
keep $\lambda=0.7$ and set $v=w=\mathbf 1/N$, the uniform choice standard for this application of RWK, which is degenerate only without edge labels (Appendix~\ref{app:degenerate}). Appendix~\ref{app:uw} compares it with the normal vectors used elsewhere
and with a degree-based choice. Each normalized Gram matrix feeds an SVC or an
SVR tuned by nested cross-validation. Since GVoys and MCRWK also produce
explicit features, we additionally fit a linear SVM or SVR on those, at a cost
linear in the number of graphs. The protocol is given in
Appendix~\ref{app:setup}. The exact methods give the same predictions, so we
report only Direct.

\begin{figure*}[t]
  \centering
  \includegraphics[width=\mainfigwidth]{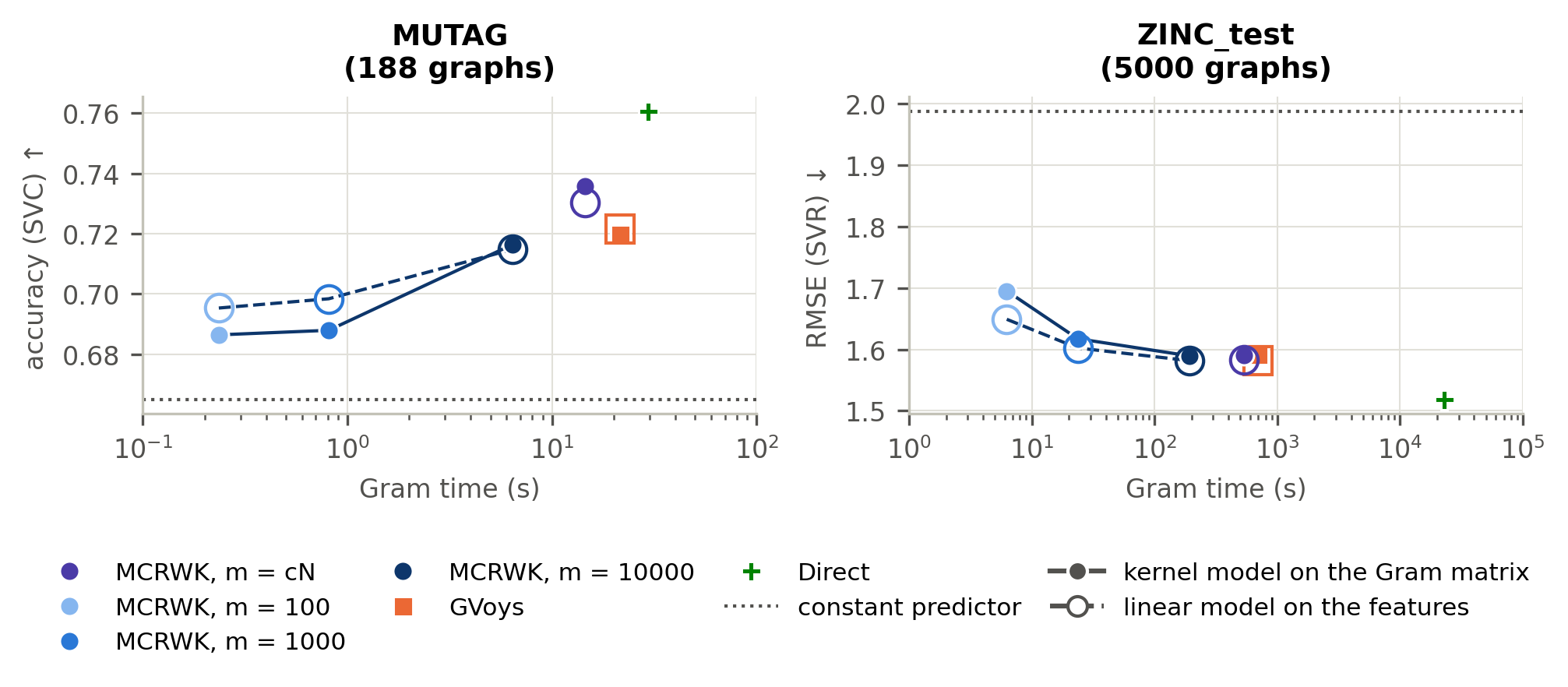}
  \caption{Prediction quality against the Gram time on labeled graphs
  ($\lambda=0.7$, uniform $v,w$)}
  \label{fig:application}
\end{figure*}

Figure~\ref{fig:application} shows the results. On ZINC\_test the exact kernel
reduces the RMSE from $1.988$ for the constant predictor, the standard deviation of the target, to $1.518$, but its
Gram matrix takes $2.3\times10^{4}$\,s, that is 6.3 hours. MCRWK with $m=10^4$
reaches $1.590$ in 191\,s, $119\times$ faster, recovering 85\% of that
improvement, and $m=10^3$ reaches $1.617$ in 24\,s. The RMSE decreases with the budget ($1.695$, $1.617$, $1.590$), and at matched runtime MCRWK is slightly ahead of
GVoys.

On MUTAG the exact kernel reaches an accuracy of $0.761$ against $0.665$ for the
majority class, and MCRWK with $m=cN$ reaches $0.736$ in 14\,s, above GVoys
($0.720$ in 21\,s). The remaining gap of about two points is consistent with the
error analysis: with four edge labels and uniform $q$, $\lambda=0.7$ exceeds
$q_{\min}=1/4$, so the variance of the labeled estimator is not controlled by
the bound of Section~\ref{sec:variance_bounds}, and Appendix~\ref{app:lambda}
shows the error growing with $\lambda$ in this regime. The linear models on the
explicit features are as good as the kernel models built on the same estimators, so MCRWK does not need the Gram matrix. Results on PTC\_MR, AIDS and the full table of
models are reported in Appendix~\ref{app:tu}.

\section{Discussion}
The experiments support the main claim of the paper: MCRWK is a faster and more accurate estimator for RWKs, with a per-pair cost independent of graph size and, at matched runtime, a lower approximation error than GVoys. The gain is
most visible on large sparse graphs, where MCRWK avoids auxiliary random
variables, anchor points, and product-graph construction.

\subsection{Limitations}\label{sec:limitations}

\paragraph{Choice of $q$.}
In the labeled case the variance depends on the proposal distribution $q$.
Our bound depends on $q$ only through $R(q)$, which is minimized by the
uniform distribution. This is because the bound uses
$\hat\phi_i \leq w_\infty$ and ignores that $\phi_i(\mathbf i)$ is small
for rare label sequences. The exact second moment depends on
$\sum_{\mathbf i} \phi_1(\mathbf i)^2 \phi_2(\mathbf i)^2 / q(\mathbf i)$,
so proposals based on empirical label frequencies can reduce the variance.
Since label sequences are shared across the dataset, $q$ must be chosen
globally. The choice of $q$ also affects runtime: with uniform $q$ on
imbalanced labels, many walks stop early with zero weight. 
The effects of $q$ on runtime and estimation error are studied in Appendix~\ref{ap:q_sampling}; adaptive choices of $q$ are left for future work.
The bound~\eqref{var-labeled} is also finite only when $\lambda<q_{\min}$, so
the $\lambda=0.7$ of our experiments lies outside that regime: the estimator
stays unbiased and its error still falls with $m$, but more slowly than
$m^{-1/2}$ (Appendices~\ref{ap:convergence} and~\ref{app:lambda}).

\paragraph{Coefficient sequences.}
MCRWK requires $C = \sum_k \mu_k < \infty$ to sample the walk length $K$. With
normalized matrices this is mild: $P_s$ is stochastic, so $v^\top A_\times^k w$
typically converges to a positive constant, and for non-summable $\mu_k$ the
kernel itself is infinite. But finite $C$ is not enough for efficiency: a walk
takes $O(K)$ steps, so the expected cost is $O(m\,\mathbb{E}[K])$ with
$\mathbb{E}[K] = \sum_k k\mu_k / C$. For heavy-tailed sequences, such as
$\mu_k \propto 1/k^2$, $C$ is finite but $\mathbb{E}[K]$ is infinite.
The estimator is still unbiased, but its expected runtime is unbounded.
Thus, MCRWK is efficient when $\sum_k k\mu_k < \infty$, which holds for
the geometric and exponential kernels.

\subsection{Why random walk kernels}\label{sec:usage}

Although GNNs are now the default in graph learning, RWKs remain useful
\citep{GraphKernelsSotaBogwardt2020}. They require no training and
lead to convex problems such as SVM or kernel ridge regression. This
makes them attractive for small datasets, where GNNs often overfit and
simple baselines are competitive \citep{errica2020fair}. RWKs are also interpretable, since the kernel counts common walks, and can be
used in Gaussian processes to estimate uncertainty. Finally,
random walk statistics are widely used inside modern models: as
positional encodings in GNNs and graph transformers
\citep{dwivedi2022graph, rampasek2022recipe}, as learnable kernels
\citep{rwk-for-kcn}, and as graph random features
\citep{grf-choromanski23a, general-grf, grf-swing-2026}.

MCRWK is therefore most useful for repeated similarity queries and for building
a full Gram matrix, as a feature-engineering step in pipelines where the dataset
is too small to train a GNN.

\section{Conclusion}

We introduced Monte Carlo Random Walk Kernel (MCRWK), an estimator for random walk graph kernels with $O(m)$-time scaling. The method rewrites the RWK as an expectation over a shared random walk length sampled directly from the normalized kernel coefficient distribution. This yields an unbiased estimator that avoids explicit product-graph construction, auxiliary Rademacher variables, and anchor-point selection.

We derived MCRWK for unlabeled and labeled graphs, showed how it recovers geometric and exponential RWK special cases, and provided variance bounds demonstrating that the sample budget required for a target precision is independent of graph size. Empirically, MCRWK is faster than GVoys and classical RWK baselines on large graphs, and more accurate than GVoys at matched runtime.

Future work includes improving the label-sequence proposal
distribution in the labeled case and extending the method
to richer graph attributes.

\section*{AI Use Statement}
In this work, we used generative AI tools to identify and summarize relevant literature, to suggest revisions to the algorithms and experimental procedures, and to proofread and rephrase parts of the manuscript. The authors reviewed all AI-assisted suggestions, checked the cited works against the original sources, and verified the final methods and experimental results. We take full responsibility for the content of this submission.

\bibliographystyle{plainnat}
\bibliography{references}

\clearpage
\appendix
\begin{center}{\Large\bfseries Monte Carlo Random Walk Kernel: Supplementary Materials}\end{center}
\section{Appendix}

\subsection{Algorithm for unlabeled case}
\label{ap:algorithm}

\begin{breakablealgorithm}
\label{alg:mcrwk_unlabeled}
\caption{Monte Carlo estimation of the Gram matrix $K \in \mathbb{R}^{N_G \times N_G}$ --- unlabeled case}
\begin{algorithmic}[1]

\Require Normalized transition matrices $P_1, \dots, P_{N_G}$, initial distributions
$v_1, \dots, v_{N_G}$, stopping distributions $w_1, \dots, w_{N_G}$,
sequence $\mu_k$, sum of sequence $C$, number of samples $m$

\Ensure Gram matrix $\hat{K}$ where $\hat{K}_{ij} \approx k(G_i, G_j)$

\State Sample shared walk lengths $K_1, \dots, K_m \sim \mathbb{P}(K = k) = \dfrac{\mu_k}{C}$

\For{$i = 1, \dots, N_G$}
    \For{$s = 1, \dots, m$}
        \State Sample $x_0^{(i,s)} \sim v_i$
        \For{$j = 1, \dots, K_s$}
            \State $x_j^{(i,s)} \sim P_i(x_{j-1}^{(i,s)}, \cdot)$
        \EndFor
        \State $F_{i,s} \gets w_i\!\left(x_{K_s}^{(i,s)}\right)$
    \EndFor
\EndFor

\For{$i = 1, \dots, N_G$}
    \For{$j = 1, \dots, i$}
        \State $\hat{K}_{ij} \gets \hat{K}_{ji} \gets \dfrac{C}{m}\displaystyle\sum_{s=1}^{m} F_{i,s} \cdot F_{j,s}$
    \EndFor
\EndFor

\State \Return $\hat{K}$

\end{algorithmic}
\end{breakablealgorithm}

\subsection{Degenerate kernel for uniform boundary vectors}
\label{app:degenerate}

\cite{VishwanathanEfficientKernel2010} use uniform distributions for the
vectors $v$ and $w$. Without edge labels this choice makes the kernel
degenerate. The transition matrix $P_i=D_i^{-1}A_i$ is row-stochastic, so
$P_i^K w_i = w_i$ for a uniform $w_i$, and
\[
v_i^\top P_i^K w_i = v_i^\top w_i = \frac{1}{n_i},
\qquad
k(G_i,G_j) = \frac{C}{n_i n_j},
\]
where $n_i$ is the number of vertices of $G_i$: the kernel sees only the graph
sizes. The unlabeled experiments therefore use random $v$ and $w$, a normalized
degree distribution works as well.

With edge labels the degeneracy disappears. The label-specific matrices
$P_i^{(l)}$ are sub-stochastic, since each row of $P_i^{(l)}$ sums to the mass
$Z_i^{(l)}(x)\le 1$ of label-$l$ edges at $x$ rather than to one, so a product
$P_i^{(l_1)}\cdots P_i^{(l_K)}$ applied to a uniform $w_i$ still depends on the
labeled structure of the graph. This is why the application experiments of
Section~\ref{sec:exp-app} can use uniform boundary vectors, and
Appendix~\ref{app:uw} compares them with the random ones.

\subsection{Experimental setup}
\label{app:setup}
\label{app:gvoys-implementation}

\paragraph{Baselines.}
Direct builds the product graph explicitly, so we run it only for $N<128$,
and Sylvester, which applies to unlabeled graphs only, for $N\le 512$. CG and FPI solve the
linear system matrix-free through the Kronecker structure, without forming the
product graph, to a relative residual of $10^{-10}$. The reference value $k$ is
computed by Direct for $N<128$ and by CG otherwise.

\paragraph{GVoys. }
GVoys was introduced for symmetric matrices such as the adjacency matrix $A$.
Each random feature joins a walk started from $v$ and a walk started from $w$ at
a common anchor vertex, with the second walk traversed backwards. Our kernel is
defined on the row-normalized, non-symmetric matrix $P=D^{-1}A$. To apply GVoys
to the same kernel as the other methods, we use the symmetric similarity
transform
\[
S=D^{-1/2}AD^{-1/2}=D^{1/2}PD^{-1/2}
\]
and transformed boundary vectors
\[
v'=D^{-1/2}v,\qquad w'=D^{1/2}w.
\]
These satisfy $v'^\top S^kw'=v^\top P^kw$ for every $k$. With edge labels, each
$P^{(l)}$ is transformed using the same $D$. We made two corrections to the reference implementation. First, neighbours are
sampled uniformly, so the load of a step $x\to y$ is multiplied by
$\deg(x)S_{xy}$. Second, the modulation function must satisfy
$\sum_{j=0}^{k}f_jf_{k-j}=\mu_k$, which for the geometric kernel gives
$f_k=\binom{2k}{k}(\lambda/4)^k$ rather than $\lambda^k$, and for the
exponential kernel $f_k=(\lambda/2)^k/k!$. Random numbers are drawn per feature from seeded streams. We use halting
probability $p_{\text{halt}}=0.2$ and all vertices as anchors, with 100 outer
samples in the scaling experiment and 200 in the Gram-matrix and application
experiments.

\paragraph{MCRWK.}
Each graph uses two independent walk replicas: their average gives the
off-diagonal features and their product the unbiased diagonal. For the linear
budget $m=cN$ the constant $c$ is calibrated once per setting so that MCRWK and
GVoys take the same time: at $N=1024$ in the scaling experiment, and on 50
random graphs per dataset, with $N$ the mean graph size, in the Gram-matrix and
application experiments.

\paragraph{Synthetic graphs.}
The synthetic experiments (Section~\ref{sec:scaling-with-graph-size} and
Appendices~\ref{ap:synthetic}, \ref{ap:convergence} and~\ref{app:lambda}) average
over four random-graph families: Barabasi--Albert graphs with attachment
parameter~2 and Erdos--Renyi graphs with $p=2/N$, both sparse, with mean degree
about 4 and 2, and two denser families, Watts--Strogatz small worlds with
$k=2\lfloor N/20\rfloor$ neighbours and rewiring probability $0.1$, and a
two-block stochastic block model with $p_{\text{in}}=0.15$ and $p_{\text{out}}=0.02$.
The exact iterative solvers are slowest on the dense families, so the speedups of
Section~\ref{sec:scaling-with-graph-size} are averages over sparse and dense
graphs; on the sparse Barabasi--Albert graphs alone, MCRWK with $m=10^4$ is
still $160\times$ faster than CG at $N=8192$. The error of MCRWK does not
depend on $N$ in any of the four families.

\paragraph{Protocols.}
The scaling experiment averages over at least 37 repetitions per size, each
drawing new graphs, new vectors $v,w$ and new random seeds. The Gram-matrix and
application experiments use the full datasets. In the application experiment we
normalize each Gram matrix to unit diagonal, select the penalty
$C\in\{10^{-3},\dots,10^{2}\}$ and, for SVR, $\epsilon\in\{0.01,0.1,0.5\}$ by
nested cross-validation with 5 outer and 3 inner folds, standardize the
regression targets within each fold, and average over three seeds. On MUTAG,
Direct takes 29\,s against 115\,s for FPI and 144\,s for CG, which is why we
report Direct as the exact baseline. The linear models are a linear SVM with
squared hinge loss and a linear SVR, for MCRWK the features are the averaged
replicas, whose inner products form the PSD Gram matrix with the biased
diagonal.

\paragraph{Implementation.}
All methods are implemented in Python with NumPy and SciPy and run on a single
CPU thread of an Intel Xeon Gold 6248R at 3.00\,GHz with 768\,GB of RAM. Code
and scripts for all experiments are in the repository linked in the abstract.

\subsection{Variance analysis for labeled graphs}
\label{app:labeled-variance}

In the labeled case the feature is
$F_r^{(p,s)}=\hat\phi_r(L^{(p,s)})/\sqrt{q(L^{(p,s)})}$,
where $\hat\phi_r$ is the output of the Walk function.
Let
\[
Z_r^{(l)}(x)=\sum_y P_r^{(l)}[x,y]
\]
be the mass of label $l$ edges at vertex $x$.
Since $P_r^{(l)}$ is normalized by the full degree,
$\sum_l Z_r^{(l)}(x)\leq1$.
For a label sequence $\mathbf i\in[d]^k$, the walk returns
\[
\hat\phi_r(\mathbf i)
=w_r(x_k)\prod_{t=1}^{k}Z_r^{(i_t)}(x_{t-1}).
\]
Write $q_{\min}=\min_l q_l$ and
$R(q)=\mathbb E_K[q_{\min}^{-K}]$.

\paragraph{Variance decomposition.}
The $mn$ feature products are not i.i.d.: products with the
same index $p$ share the walk length $K_p$.
Conditional on $K_p$, the label sequences and the corresponding
walks are sampled independently.
Writing $Y=F_1^{(1,1)}F_2^{(1,1)}$, the law of total variance gives
\begin{equation}
\begin{aligned}
\mathrm{Var}(\hat k)
&=\frac{C^2}{mn}\,
  \mathbb E[\mathrm{Var}(Y\mid K_1)] +\frac{C^2}{m}\,
  \mathrm{Var}(\mathbb E[Y\mid K_1]).
\end{aligned}
\label{var-labeled-decomp}
\end{equation}

\paragraph{First term.}
Fix $K_1=k$.
Given the labels, the walks on $G_1$ and $G_2$ are independent,
and $q(\mathbf i)\geq q_{\min}^k$.
Hence
\[
\begin{aligned}
\mathbb E[Y^2\mid K_1=k]
&=\sum_{\mathbf i\in[d]^k}
  \frac{\mathbb E[\hat\phi_1(\mathbf i)^2]\,
        \mathbb E[\hat\phi_2(\mathbf i)^2]}
       {q(\mathbf i)} \leq w_\infty^4q_{\min}^{-k}
  \sum_{\mathbf i\in[d]^k}
  a_1(\mathbf i)a_2(\mathbf i),
\end{aligned}
\]
where
\[
a_r(\mathbf i)
=\mathbb E\!\left[
  \prod_{t=1}^{k}Z_r^{(i_t)}(x_{t-1})^2
\right]\leq1.
\]
The expectation is over the walk for the fixed label sequence
$\mathbf i$.
Since $\sum_l Z_r^{(l)}(x)^2\leq1$ at every step,
induction on $k$ gives
$\sum_{\mathbf i\in[d]^k}a_r(\mathbf i)\leq1$.
Thus
\[
\mathbb E[Y^2\mid K_1]
\leq w_\infty^4q_{\min}^{-K_1},
\qquad
\mathbb E[\mathrm{Var}(Y\mid K_1)]
\leq w_\infty^4R(q).
\]

\paragraph{Second term.}
For any $k$,
\[
\begin{aligned}
\mathbb E[Y\mid K_1=k]
&=\sum_{\mathbf i\in[d]^k}
  \phi_1(\mathbf i)\phi_2(\mathbf i)
&\leq\prod_{r=1}^{2}
  \sum_{\mathbf i\in[d]^k}\phi_r(\mathbf i)
&=\prod_{r=1}^{2}v_r^\top P_r^kw_r
\leq w_\infty^2,
\end{aligned}
\]
where we used $\sum_lP_r^{(l)}=P_r$.
Since $\mathbb E[Y\mid K_1]\in[0,w_\infty^2]$,
Popoviciu's inequality gives
\[
\mathrm{Var}(\mathbb E[Y\mid K_1])
\leq\frac{w_\infty^4}{4}.
\]
Substituting the two bounds into
\eqref{var-labeled-decomp} yields \eqref{var-labeled}.
Chebyshev's inequality then gives~\eqref{Chebyshev-labeled},
and hence the sufficient sample budget in
\eqref{labeled_sample_complexity}.

\paragraph{Allocation of the walk budget.}
Here $n$ controls the variance of label sampling, while $m$
controls the variance of the walk length, which cannot be
reduced by increasing $n$ alone.
For fixed $m$, increasing $n$ reduces only the first term
in \eqref{var-labeled}, while the second term $1/4$
remains unchanged.
Thus, taking $n$ larger than about $4R(q)$ gives almost
no improvement.

The cost of the estimator is proportional to the total number
of walks $B=mn$.
Substituting $m=B/n$ into \eqref{var-labeled-decomp} gives
\[
\mathrm{Var}(\hat k)
=\frac{C^2}{B}
\left(
  \mathbb E[\mathrm{Var}(Y\mid K_1)]
  +n\,\mathrm{Var}(\mathbb E[Y\mid K_1])
\right),
\]
which grows with $n$.
Therefore, $n=1$ is optimal for a fixed budget, while $n>1$
is useful only when $m$ is limited.

\subsection{Extended results for synthetic graphs}
\label{ap:synthetic}

\begin{figure*}[t]
  \centering
  \includegraphics[width=\textwidth]{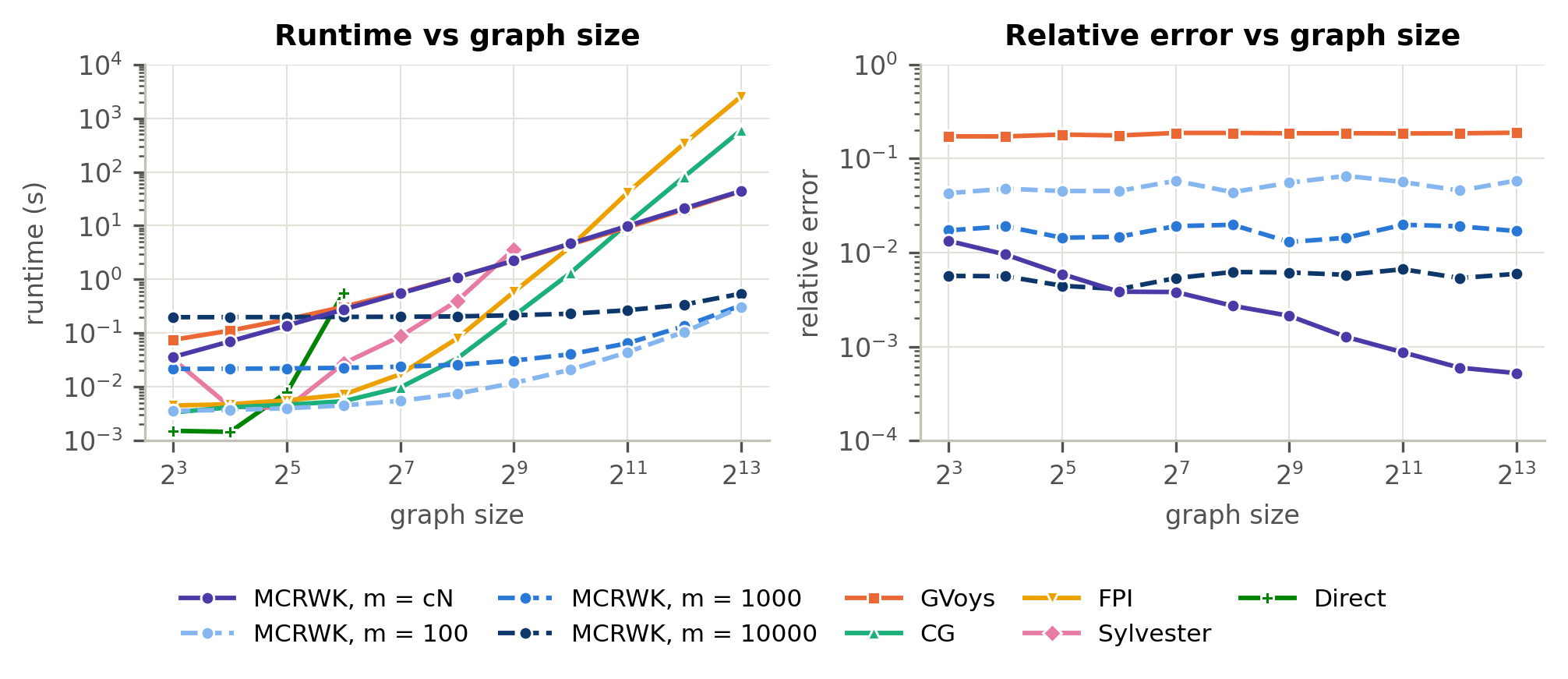}
  \caption{Runtime (left) and relative error $|\hat k-k|/k$ (right) of a single
  kernel value $k(G_1,G_2)$ versus graph size for pairs of unlabeled random graphs from the four families of Appendix~\ref{app:setup}, averaged over 30 repetitions. Direct runs for
  $N<128$, Sylvester for $N\le 512$. By construction, MCRWK with $m=cN$ and
  GVoys have matching runtimes, so their curves overlap in the left panel.}
  \label{fig:scaling_unlabeled}
\end{figure*}

Figure~\ref{fig:scaling_unlabeled} repeats the scaling experiment without edge labels, with the same graph families, sizes, methods and budgets. We run
Sylvester for $N\le 512$.

\paragraph{Runtime.}
As in the labeled case, MCRWK with a fixed budget is almost insensitive to the
graph size: at $N=8192$ it takes 0.30, 0.33 and 0.54\,s for $m=10^2,10^3,10^4$.
GVoys takes 43.5\,s and MCRWK with $m=cN$ 44.3\,s, while CG and FPI take 596\,s
and $2.5\times10^{3}$\,s. Direct reaches 0.56\,s already at $N=64$, and Sylvester 3.7\,s at
$N=512$. On small graphs CG is competitive: it is faster than the equal-time
estimators up to $N\approx 2^{11}$, and faster than MCRWK with $m=10^4$ up to
$N\approx 2^{9}$. At $N=8192$, however, MCRWK with $m=10^4$ is about
$1100\times$ faster than CG and $4700\times$ faster than FPI.

\paragraph{Accuracy.}
With a fixed budget the error of MCRWK again does not depend on the graph size:
it stays within 4.3--6.5\%, 1.3--2.0\% and 0.4--0.7\% for $m=10^2,10^3,10^4$
as $N$ grows from 8 to 8192. Without labels each tenfold increase of $m$
reduces the mean error by a factor of $3.0$, close to the
$\sqrt{10}\approx 3.2$ predicted by $\sigma(\hat k)=O(C/\sqrt m)$; the slower second step seen in the labeled case does not occur here. With the linear
budget $m=cN$ the error falls from 1.3\% at $N=8$ to 0.05\% at $N=8192$. The
gap to GVoys is larger than with labels: at matched runtime GVoys keeps an
error of 17--19\% at every size, 13 to 360 times that of MCRWK with $m=cN$.

\subsection{Gram matrix computation without labels}
\label{app:gram_time}

\begin{figure*}[t]
  \centering
  \includegraphics[width=\textwidth]{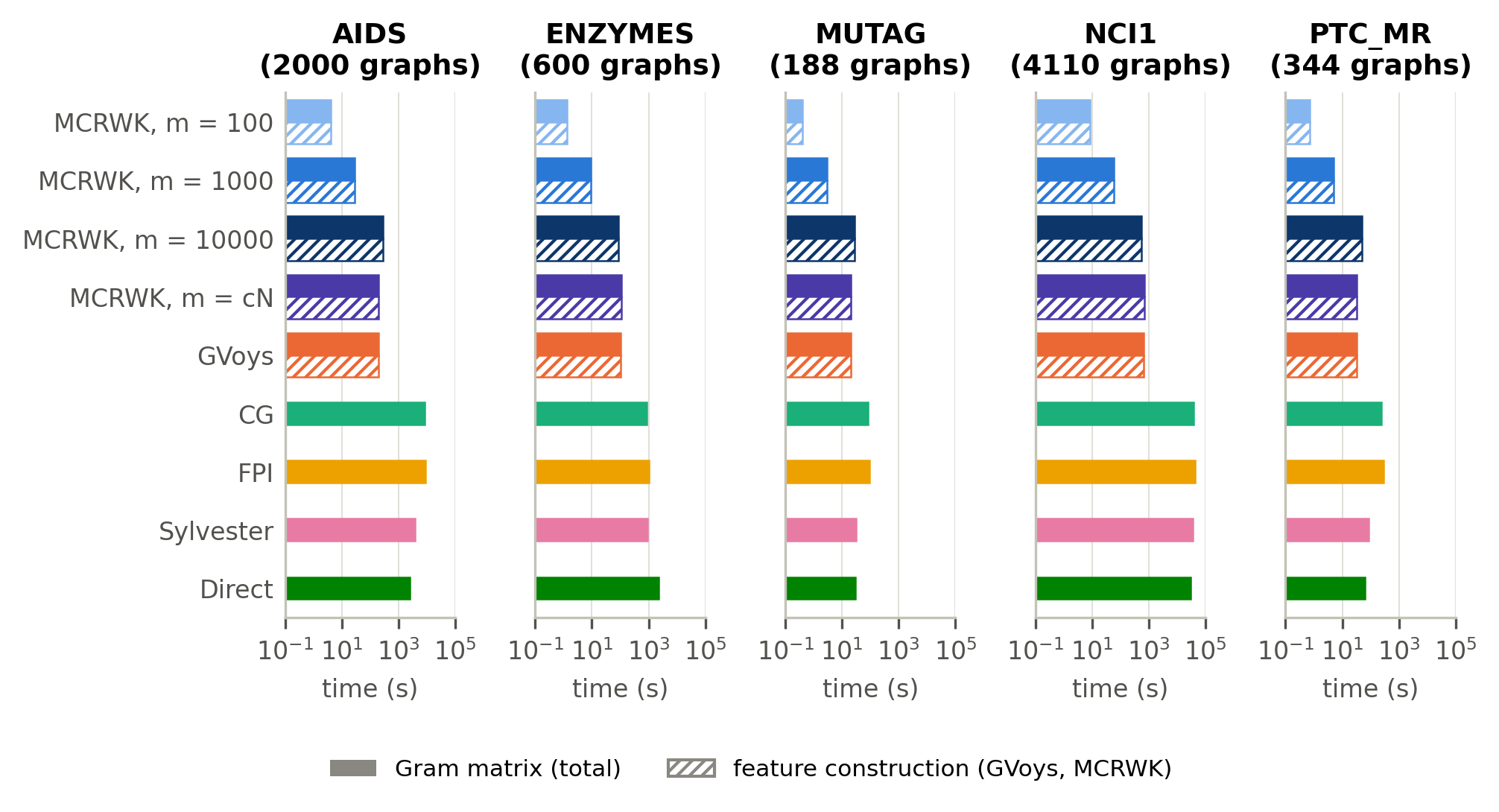}
  \caption{Time to compute the Gram matrix of a whole dataset without edge
  labels (logarithmic axis). Solid bars: total time; hatched bars: feature
  construction of GVoys and MCRWK, before the Gram product.}
  \label{fig:gram_time_unlabeled}
\end{figure*}

Figure~\ref{fig:gram_time_unlabeled} repeats the experiment of
Section~\ref{sec:exp-gram} without edge labels, on AIDS (2000 graphs), ENZYMES
(600), MUTAG (188), NCI1 (4110) and PTC\_MR (344). The setup is the same and
Sylvester is included.

The picture is the same as in the labeled case. On the largest dataset, NCI1,
all exact methods need between $3.0\times10^{4}$ and $4.2\times10^{4}$\,s (8 to 12 hours). MCRWK builds the same Gram
matrix in 8.6, 59 and 559\,s for $m=10^2,10^3,10^4$, which is 54 times faster
than Direct with $m=10^4$ and about 500 times faster with $m=10^3$. On AIDS MCRWK
takes 4.0, 29 and 277\,s. The fastest exact method there, Direct, takes
$2.4\times10^{3}$\,s, Sylvester, CG and FPI take $3.8\times10^{3}$,
$8.1\times10^{3}$ and $9.0\times10^{3}$\,s. On ENZYMES, whose graphs are larger,
CG is the fastest exact method with 858\,s, against 1.3, 9.0 and 86\,s for
MCRWK. On the two smallest datasets the advantage shrinks. On MUTAG, MCRWK with
$m=10^4$ takes 27\,s, about as long as Direct (29\,s), and on PTC\_MR it takes
49\,s against 63\,s. With $m=10^3$ MCRWK is still ten times faster on both.
GVoys and MCRWK with $m=cN$ again take similar times (640 and 678\,s on NCI1),
and feature construction accounts for more than 98\% of their total time.

The datasets also show how the two approaches scale with the number of graphs
$N_G$. From MUTAG to NCI1, $N_G$ grows 22 times and the number of graph pairs
about 480 times. Over the same range, the time of MCRWK with a fixed $m$ grows 21
times, as expected for one feature map per graph and a Gram product that
contributes little. The time of Direct grows about $10^{3}$ times, since it
solves one linear system per pair and the NCI1 graphs are also larger.

At the same budget, MCRWK is slower without labels than with them (277
against 73\,s on AIDS for $m=10^4$). In the labeled case a walk stops as soon as
the current vertex has no edge with the sampled label, so walks are shorter on
average.

\subsection{Convergence and tightness of the bounds}
\label{ap:convergence}

We check the bounds of Section~\ref{sec:variance_bounds} empirically. For each size
$N\in\{16,64,256,1024\}$ we draw 8 graphs (28 pairs) from each of the four families of Appendix~\ref{app:setup},
compute the exact kernel of every pair (Direct for $N<128$, CG otherwise), and
run MCRWK with 50 independent seeds. Budgets are nested: each seed draws
$m=10^4$ samples, and the estimate for a smaller $m$ uses the first $m$ of them.
Since the samples are i.i.d., this is a valid $m$-sample estimate, and every
statistic for a given $m$ is still computed over independent seeds. We use
$\lambda=0.7$, as in the rest of the paper, and $\lambda=0.3$. With $d=3$
labels and uniform $q$, $q_{\min}=1/3$, so the labeled bound~\eqref{var-labeled} is finite
only for $\lambda=0.3$.

\begin{figure*}[t]
  \centering
  \includegraphics[width=\textwidth]{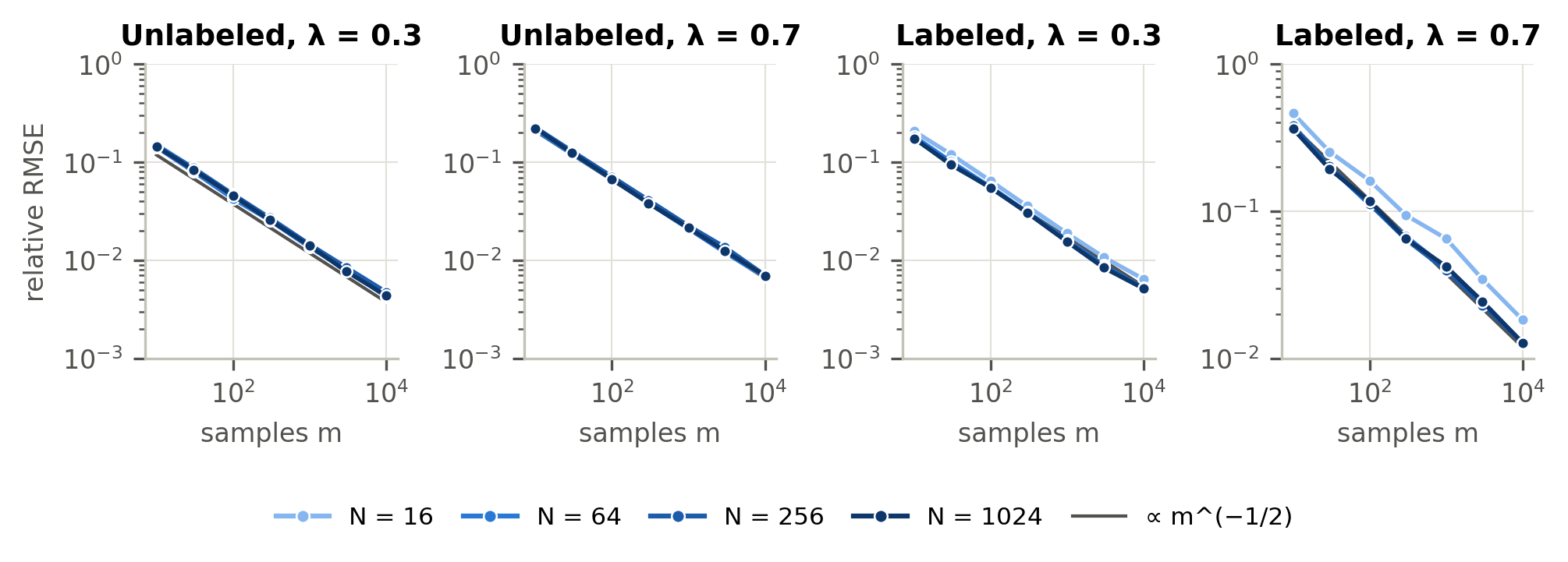}
  \caption{Relative RMSE of MCRWK over all pairs of 8 graphs versus the number
  of samples $m$, for four graph sizes. The gray line has slope $-1/2$.}
  \label{fig:convergence_rmse}
\end{figure*}

\paragraph{Rate.}
Figure~\ref{fig:convergence_rmse} shows the relative RMSE over all pairs. On
log-log axes the curves are straight lines with fitted slopes between $-0.49$
and $-0.52$ in the unlabeled case and between $-0.51$ and $-0.53$ in the labeled
case at $\lambda=0.3$, matching the $m^{-1/2}$ rate. The curves for different
$N$ coincide: the sample complexity does not depend on the graph size. In the
labeled case at $\lambda=0.7$, where the bound is infinite, the slope is
flatter, between $-0.45$ and $-0.48$: the heavy-tailed importance weights slow
the rate down, though the error still decreases with $m$.

\begin{figure}[t]
  \centering
  \includegraphics[width=\linewidth]{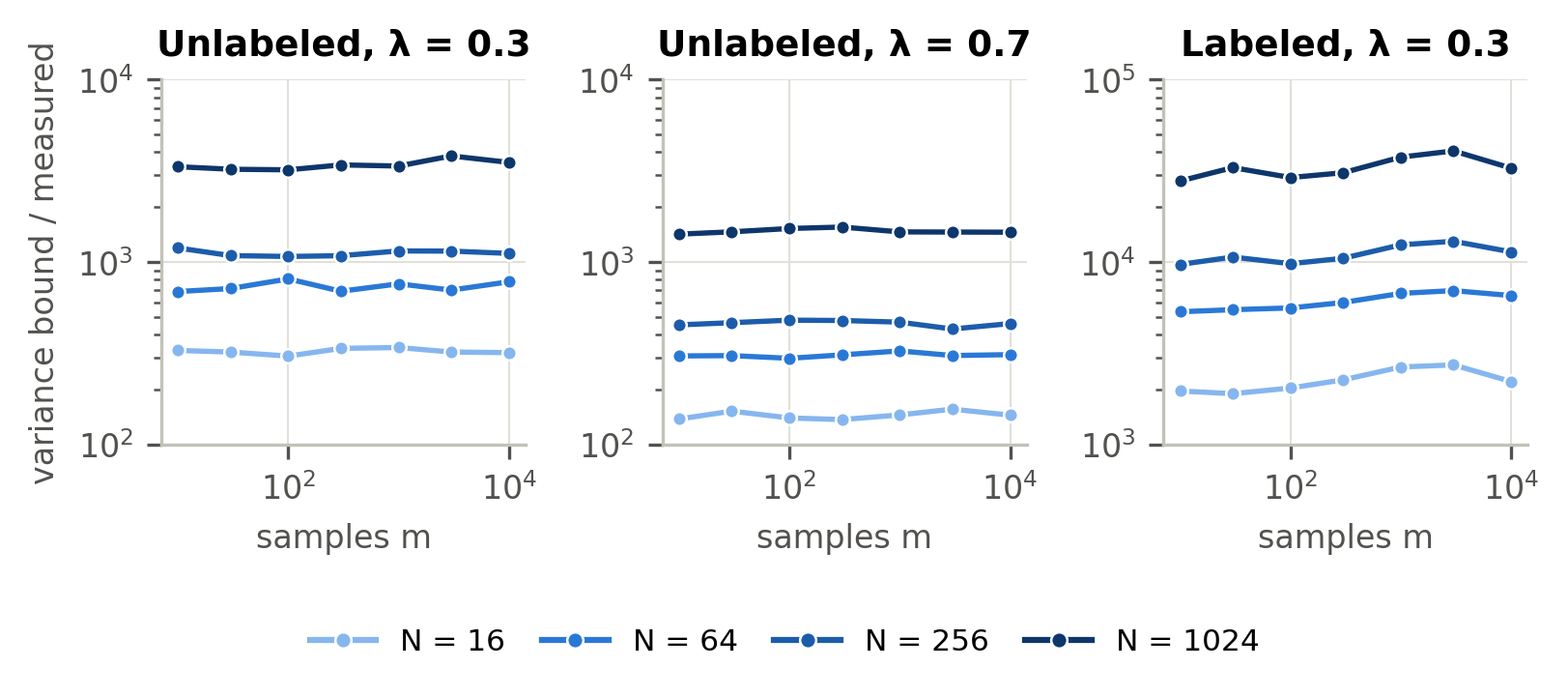}
  \caption{Variance bound~\eqref{var-unlabeled} or~\eqref{var-labeled} divided by the
  measured variance, median
  over pairs. The bound is infinite for labeled graphs at $\lambda=0.7$.}
  \label{fig:convergence_bound}
\end{figure}

\paragraph{Tightness.}
The variance bounds~\eqref{var-unlabeled} and~\eqref{var-labeled} have the right $1/m$ dependence but a loose
constant. Figure~\ref{fig:convergence_bound} shows the ratio of the bound to
the measured variance, median over pairs. It does not depend on $m$, but it
ranges from about $150$ to $3.4\times10^{3}$ in the unlabeled case and from
$2.7\times10^{3}$ to $3.8\times10^{4}$ in the labeled case, and it grows with
$N$. The bounds use the largest stopping weight $w_\infty$, whereas a typical
walk ends at a vertex of average weight. With our random $v$ and $w$ the
ratio of the two grows slowly with $N$.

\begin{figure*}[t]
  \centering
  \includegraphics[width=\textwidth]{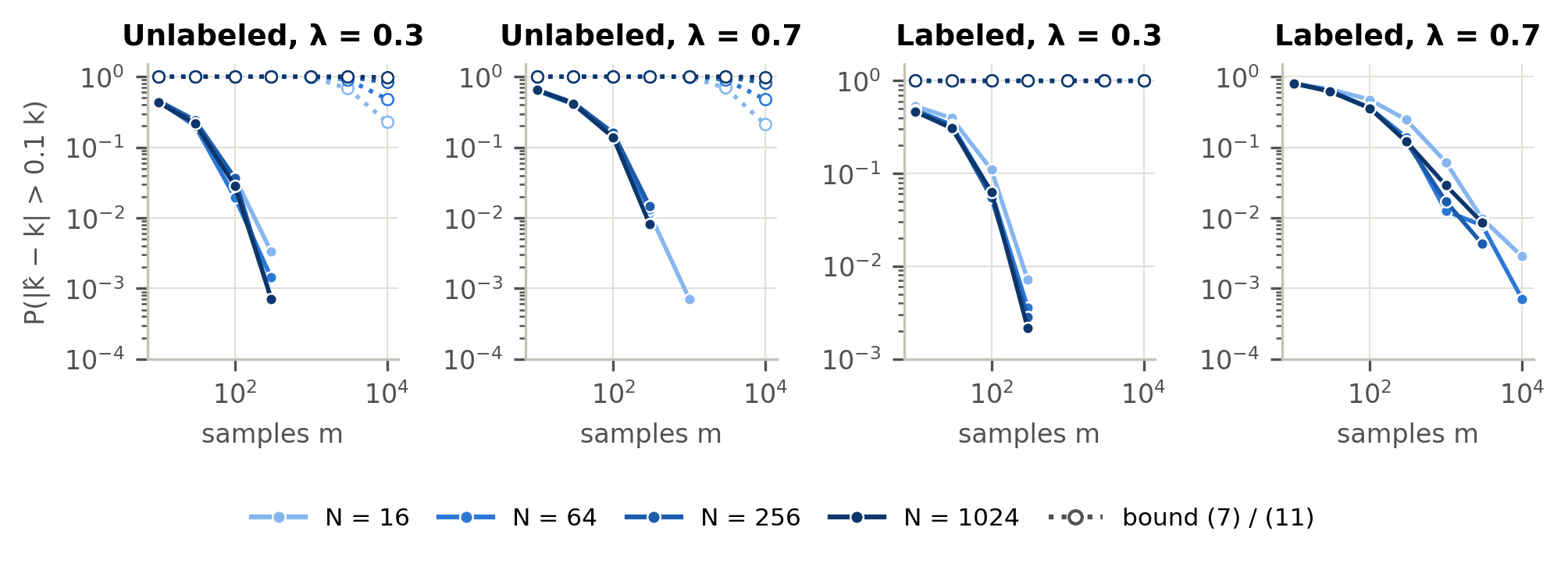}
  \caption{Frequency of a relative error above $10\%$ (solid) and its bound
  (dotted): Hoeffding~\eqref{Hoeffding-unlabeled} without labels,
  Chebyshev~\eqref{Chebyshev-labeled} with labels. For
  labeled graphs at $\lambda=0.7$ the bound is infinite and is not shown.
  Missing points have frequency $0$.}
  \label{fig:convergence_tails}
\end{figure*}

The tail bounds are correspondingly conservative
(Figure~\ref{fig:convergence_tails}). Hoeffding's bound~\eqref{Hoeffding-unlabeled} for a 10\% error
stays above $0.2$ even at $m=10^4$, while the measured frequency of such
errors falls below $10^{-2}$ already at $m=300$. Table~\ref{tab:required_m}
compares the $m$ that guarantees a relative error of at most $\varepsilon$ with
probability $1-\delta$ according to~\eqref{sampling-unlabled} and~\eqref{labeled_sample_complexity} with the smallest $m$ of our
grid that achieves it. The unlabeled bound overestimates the required budget
by two to three orders of magnitude, and the labeled bound by four to five.

\begin{table}[t]
  \centering
  \caption{Budget $m$ needed for $\Pr(|\hat k-k|>\varepsilon k)\le\delta$ with
  $\delta=0.05$, at $N=1024$: the smallest $m$ of the grid
  $\{10,30,100,\dots,10^4\}$ that achieves it, and the bound~\eqref{sampling-unlabled}
  or~\eqref{labeled_sample_complexity},
  median over pairs.}
  \label{tab:required_m}
  \begin{tabular}{llrrrr}
    \toprule
    & & \multicolumn{2}{c}{$\varepsilon=0.1$} & \multicolumn{2}{c}{$\varepsilon=0.05$} \\
    case & $\lambda$ & measured & bound & measured & bound \\
    \midrule
    unlabeled & 0.3 & 100  & $1.3\times10^{5}$ & 300  & $5.3\times10^{5}$ \\
    unlabeled & 0.7 & 300  & $1.4\times10^{5}$ & 1000 & $5.5\times10^{5}$ \\
    labeled   & 0.3 & 300  & $1.7\times10^{7}$ & 1000 & $6.7\times10^{7}$ \\
    labeled   & 0.7 & 1000 & $\infty$          & 3000 & $\infty$          \\
    \bottomrule
  \end{tabular}
\end{table}

\subsection{Choice of the label proposal $q$}
\label{ap:q_sampling}

\begin{figure*}[t]
  \centering
  \includegraphics[width=\textwidth]{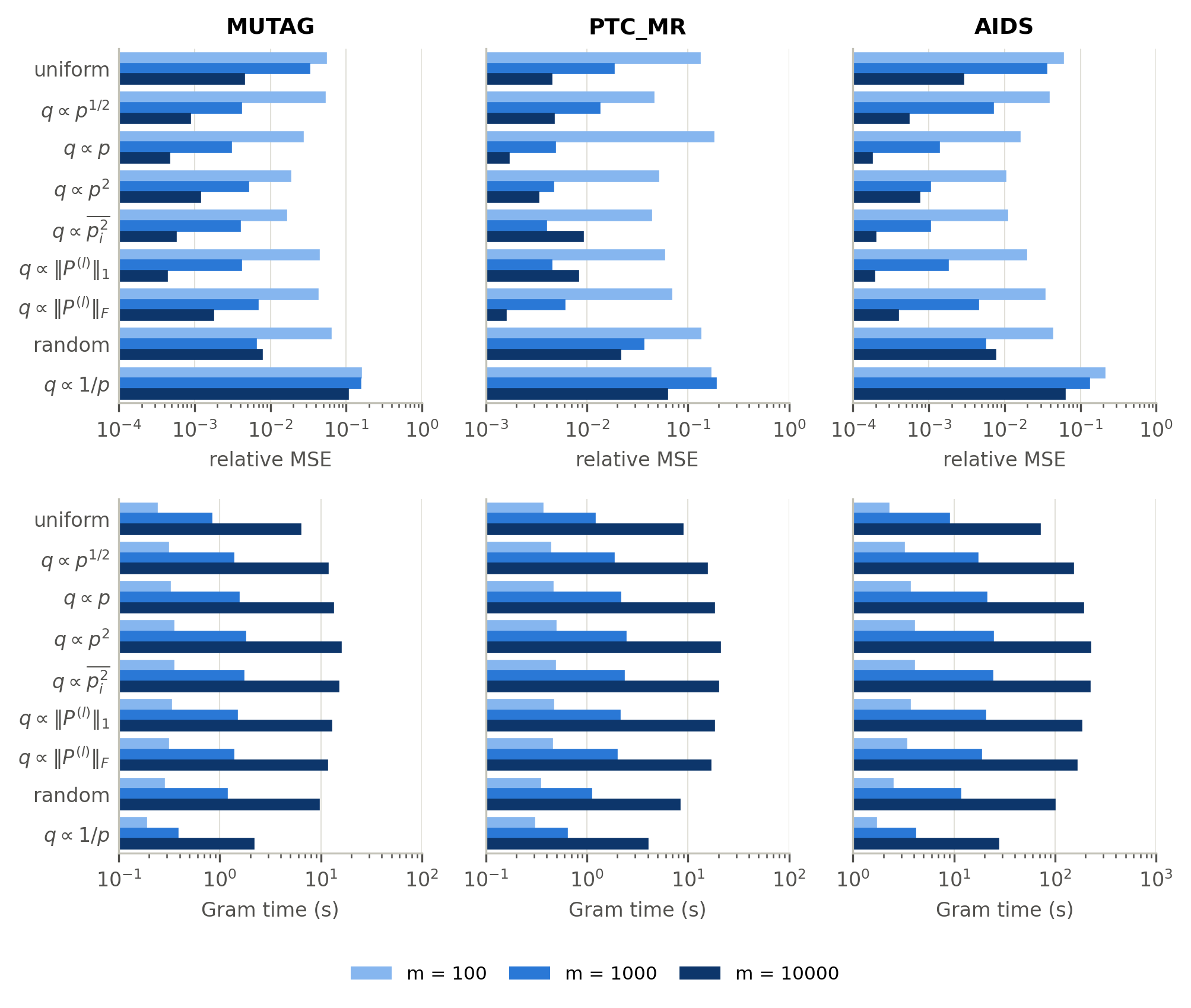}
  \caption{Relative MSE of the Gram matrix (top) and Gram time (bottom) for
  different label proposals $q$, $\lambda=0.7$, averaged over 5 seeds.}
  \label{fig:q_sampling_07}
\end{figure*}

\begin{figure*}[t]
  \centering
  \includegraphics[width=\textwidth]{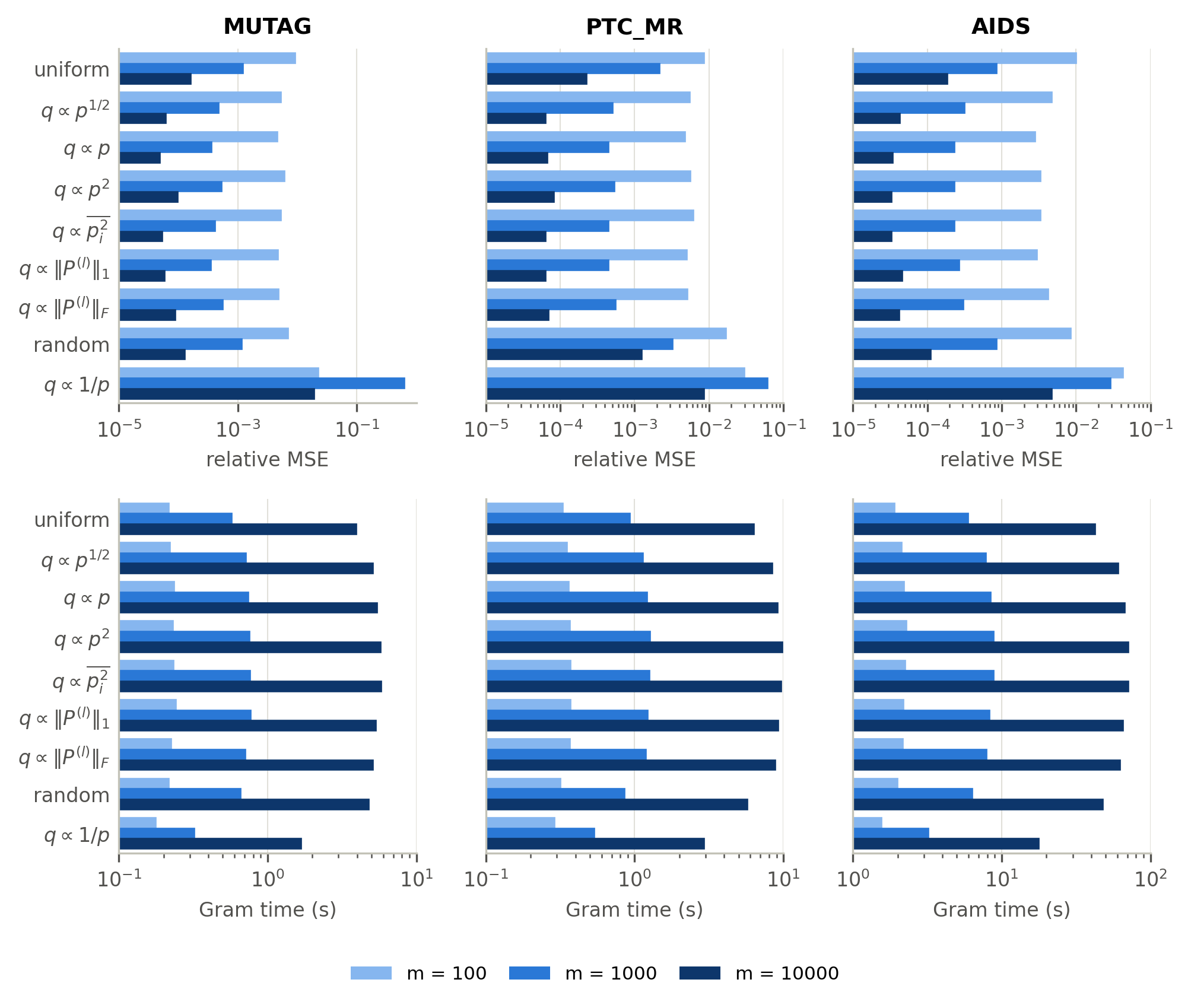}
  \caption{Same as Figure~\ref{fig:q_sampling_07} for $\lambda=0.3$.}
  \label{fig:q_sampling_03}
\end{figure*}

In the labeled case the variance depends on the proposal $q$ through the
importance weights (Section~\ref{sec:variance_bounds}). We compare nine proposals on MUTAG, PTC\_MR
and AIDS with their edge labels. Let $p_l$ be the fraction of edges with label
$l$ in the dataset and $p_{il}$ that fraction in graph $G_i$. The proposals are
uniform; $q_l\propto p_l^{\alpha}$ for $\alpha\in\{1/2,1,2\}$; $q_l\propto$ the
average of $p_{il}^2$ over the graphs; $q_l\propto\sum_i\|P_i^{(l)}\|_1$ and
$q_l\propto\sum_i\|P_i^{(l)}\|_F$; a random $q$; and $q_l\propto 1/p_l$ as a
deliberately poor control. Under a mean-field approximation, where at every
vertex the mass of label $l$ is $p_l$, the second moment per step is
$\sum_l p_l^4/q_l$, minimized by $q\propto p^2$. We use
$m\in\{10^2,10^3,10^4\}$, $n=1$, $\lambda\in\{0.3,0.7\}$ and 5 seeds, shared by
all proposals. We report the relative MSE of the Gram entries against Direct,
averaged over the seeds, and the Gram time
(Figures~\ref{fig:q_sampling_07} and~\ref{fig:q_sampling_03}).

The uniform proposal is never the most accurate. At $m=10^4$ and
$\lambda=0.7$, proposals that follow the label frequencies ($q\propto p$,
$q\propto\overline{p_i^2}$, $\|P^{(l)}\|_1$) reduce the relative MSE by a factor
of 10 on MUTAG and 16 on AIDS. On PTC\_MR only $q\propto p$ helps, by a factor
of 3, while $q\propto\overline{p_i^2}$ and $\|P^{(l)}\|_1$ are about twice worse
than uniform. At $\lambda=0.3$ every frequency-based proposal gains a factor of
3--6 on all three datasets. The inverse proposal is one to two orders of magnitude worse, as
expected. The price is runtime. Uniform $q$ often samples labels that a graph
does not contain or that are rare at the current vertex, and such walks end
early. Frequency-based proposals keep walks alive, which makes the Gram
computation 1.4--2.7 times slower (e.g.\ $71$\,s against $192$\,s on AIDS at
$\lambda=0.7$ for $q\propto p$). Even per unit of time they remain clearly
better on MUTAG and AIDS. The mean-field optimum $q\propto p^2$ is the best proposal on AIDS at
$\lambda=0.3$, but only sixth of nine there at $\lambda=0.7$ and never the best
on MUTAG or PTC\_MR, where labels along a walk are correlated.

\subsection{Number of label sequences per length}
\label{ap:n_sampling}

\begin{figure*}[t]
  \centering
  \includegraphics[width=\textwidth]{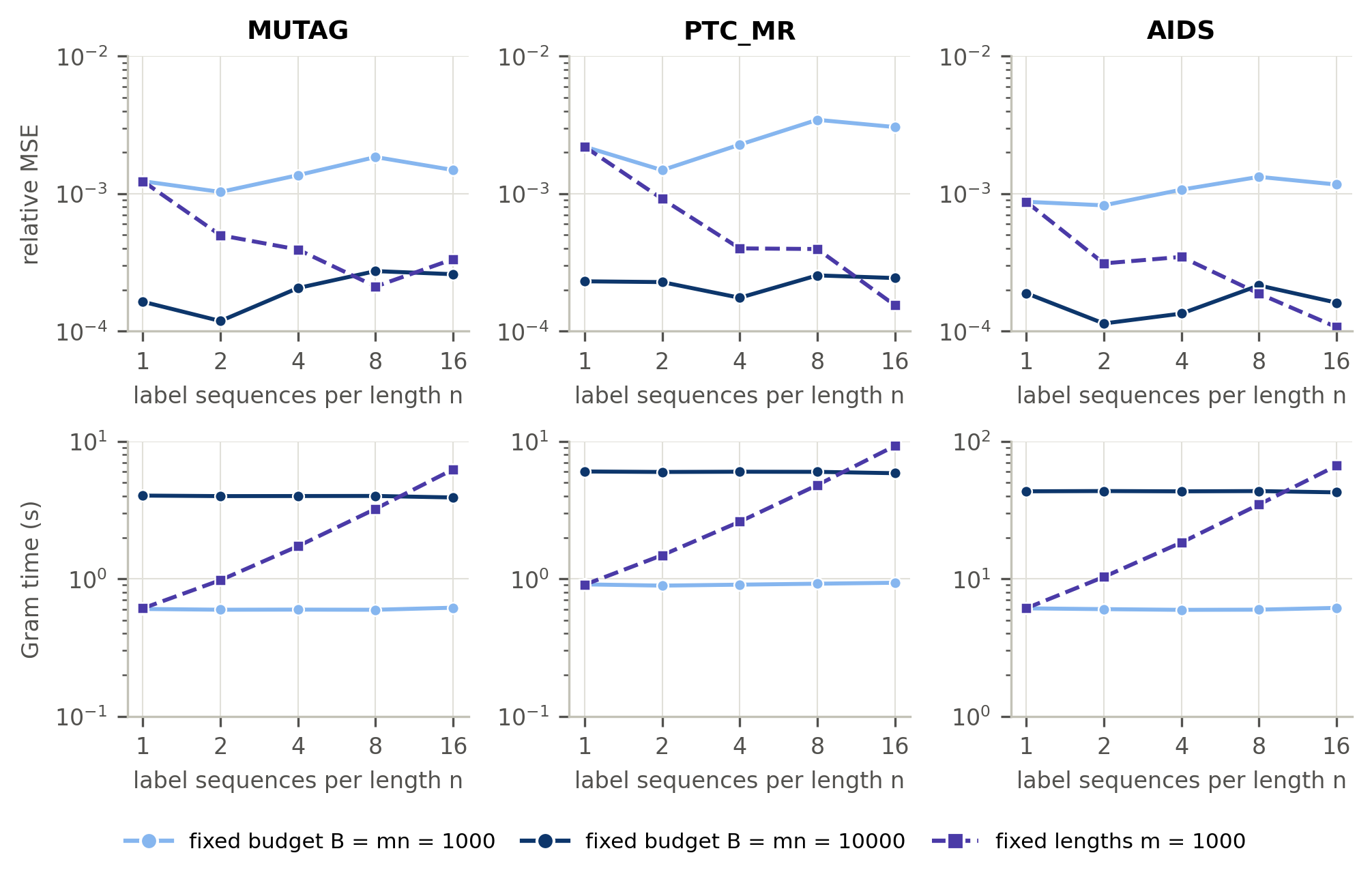}
  \caption{Relative MSE (top) and Gram time (bottom) versus the number $n$ of
  label sequences per length, for a fixed budget $B=mn$ and for a fixed number
  of lengths $m=10^3$, $\lambda=0.3$, averaged over 5 seeds.}
  \label{fig:n_sampling_03}
\end{figure*}

\begin{figure*}[t]
  \centering
  \includegraphics[width=\textwidth]{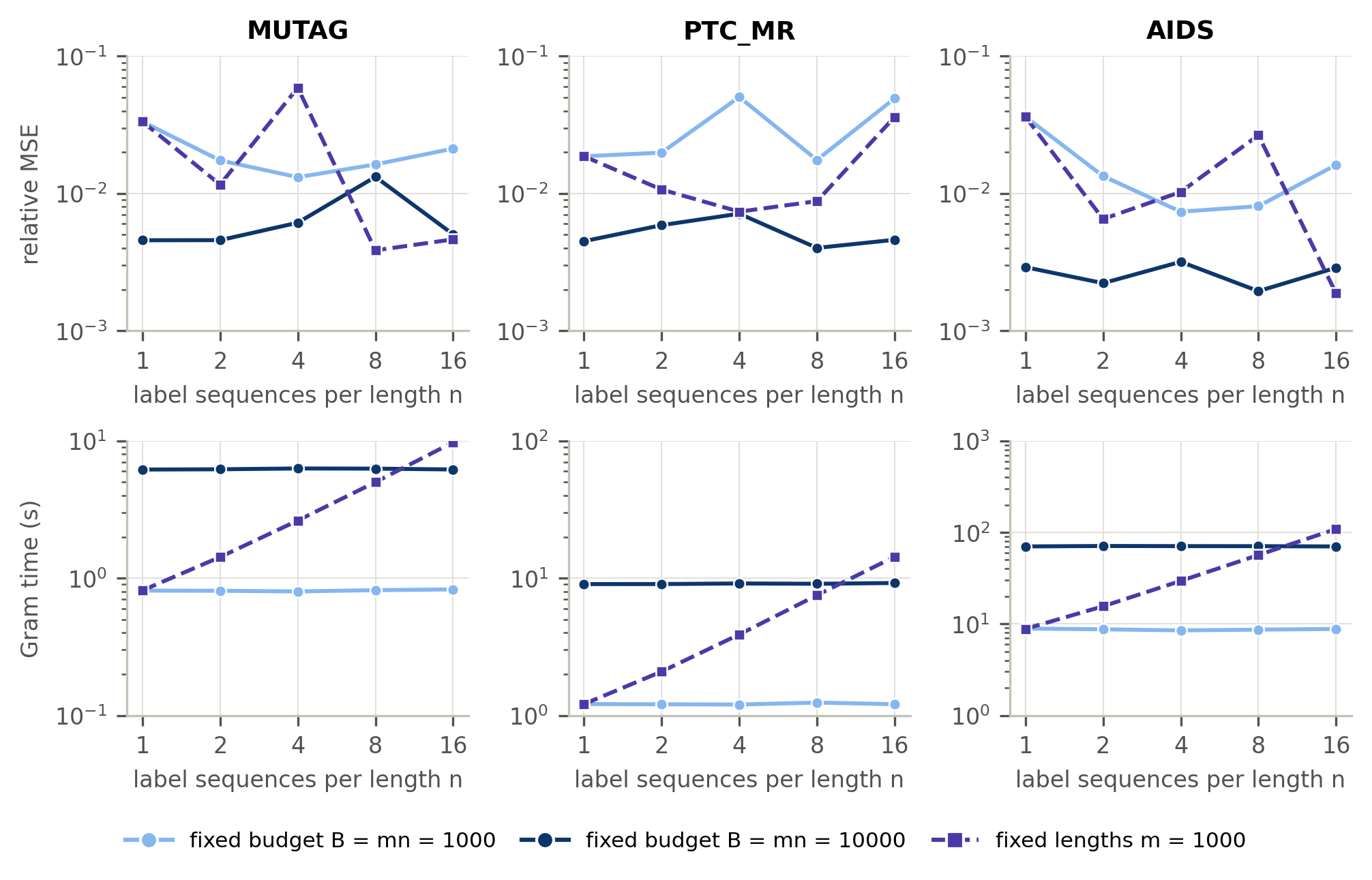}
  \caption{Same as Figure~\ref{fig:n_sampling_03} for $\lambda=0.7$.}
  \label{fig:n_sampling_07}
\end{figure*}

Section~\ref{sec:variance_bounds} predicts that for a fixed total budget $B=mn$ the variance does not
decrease with the number $n$ of label sequences per length, so that $n=1$ is
optimal. For a fixed number of lengths $m$, increasing $n$ reduces only the
label-sampling part of the variance. We test both designs on MUTAG, PTC\_MR and
AIDS with uniform $q$, $n\in\{1,2,4,8,16\}$ and 5 seeds: a fixed budget
$B\in\{10^3,10^4\}$, and a fixed $m=10^3$
(Figures~\ref{fig:n_sampling_03} and~\ref{fig:n_sampling_07}).

The runtime follows the design: it is constant in $n$ at a fixed budget and
grows linearly with $n$ at a fixed $m$ (from 8.8 to 109\,s on AIDS). At a
fixed budget we see no systematic benefit of $n>1$. At $\lambda=0.3$ and
$B=10^4$ the relative MSE stays between $1.1\times10^{-4}$ and
$2.7\times10^{-4}$ for all $n$, within the spread between seeds. At
$\lambda=0.7$ and $B=10^3$, $n=2$--$4$ even gave a lower error than $n=1$ on
MUTAG and AIDS. We attribute this to the heavy-tailed weights at
$\lambda>q_{\min}$. With a fixed $m=10^3$ and
$\lambda=0.3$ the error decreases with $n$, by a factor of 6 to 15 at $n=8$--16,
without saturating in this range. Each additional sequence, however, costs as
much as an additional length. Since increasing $m$ also reduces the
length-sampling part of the variance, $n=1$ with a larger $m$ remains the
simplest choice.

\subsection{Effect of the decay parameter $\lambda$}
\label{app:lambda}

\begin{figure*}[t]
  \centering
  \includegraphics[width=\textwidth]{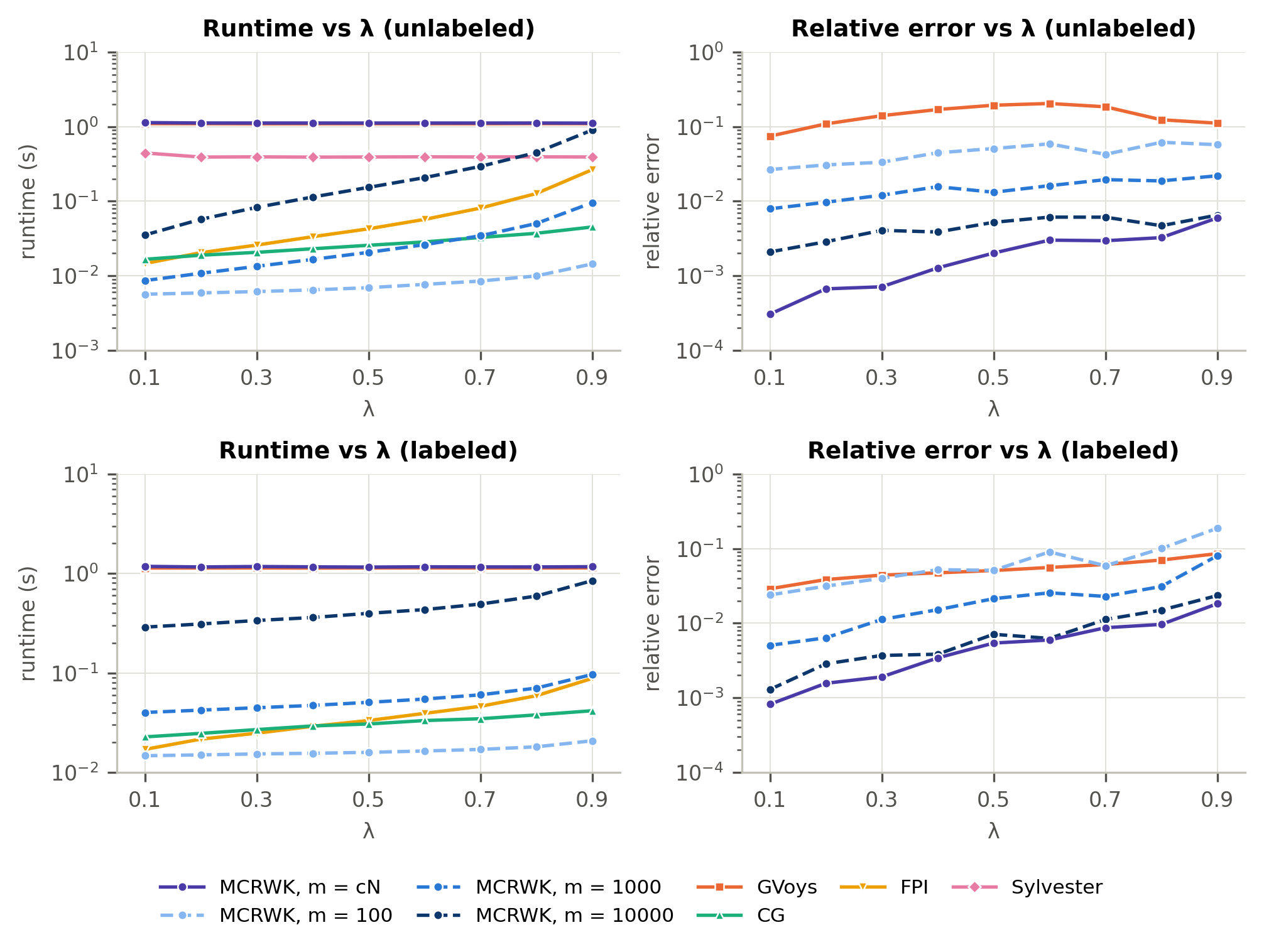}
  \caption{Runtime and relative error of $k(G_1,G_2)$ against $\lambda$ for
  pairs of synthetic graphs with $N=256$ vertices, without (top) and with
  (bottom) edge labels.}
  \label{fig:lambda_synthetic}
\end{figure*}

\begin{figure*}[t]
  \centering
  \includegraphics[width=\textwidth]{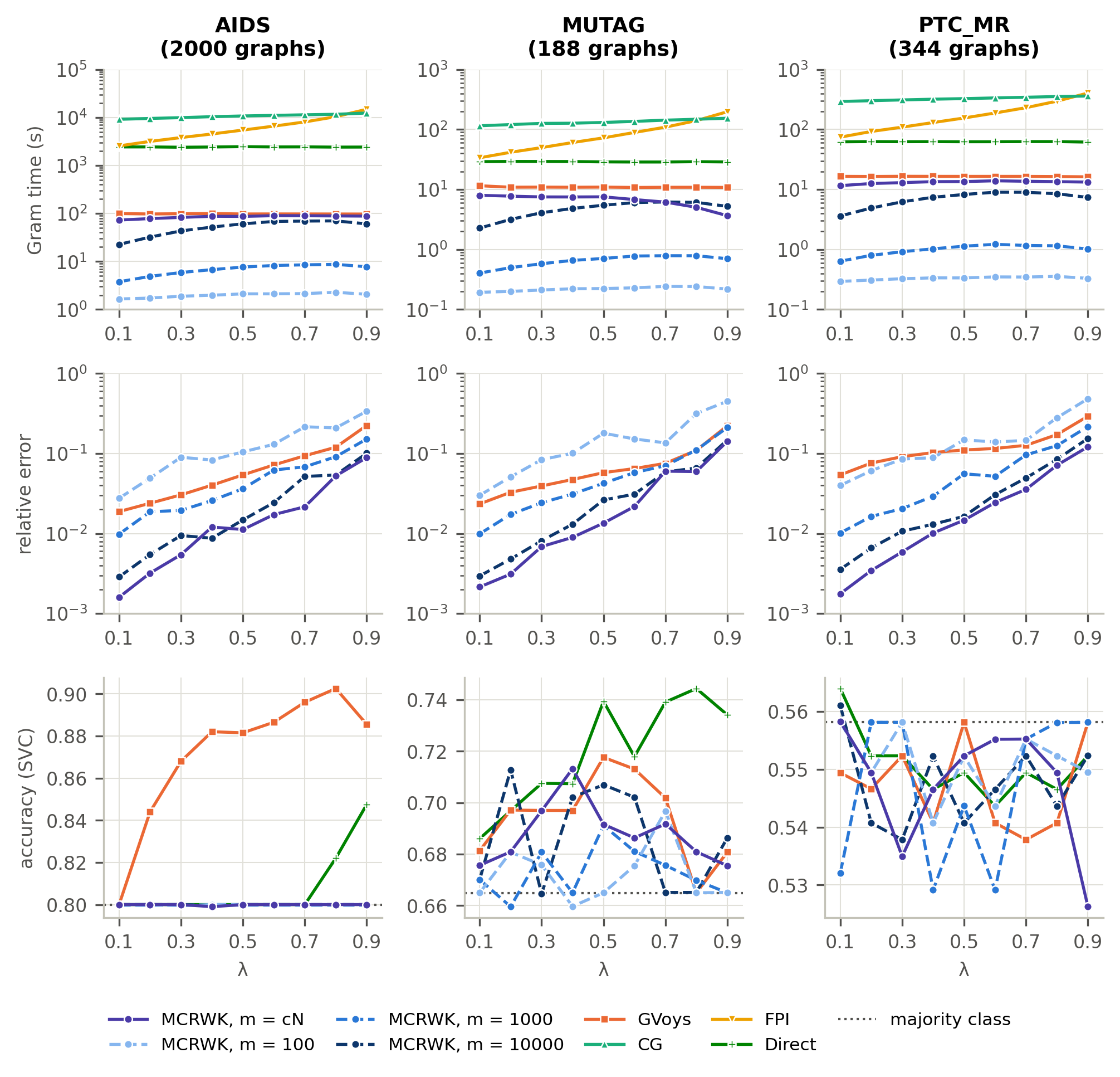}
  \caption{Gram matrix time, relative error of the off-diagonal Gram entries
  against Direct, and SVC accuracy against $\lambda$ on labeled TU datasets, median over three seeds.
  The exact methods compute the same kernel and reach the same accuracy, so
  only Direct is shown in the bottom row, the dotted line is the fraction of
  the most frequent class.}
  \label{fig:lambda_tu_labeled}
\end{figure*}

\begin{figure*}[t]
  \centering
  \includegraphics[width=\textwidth]{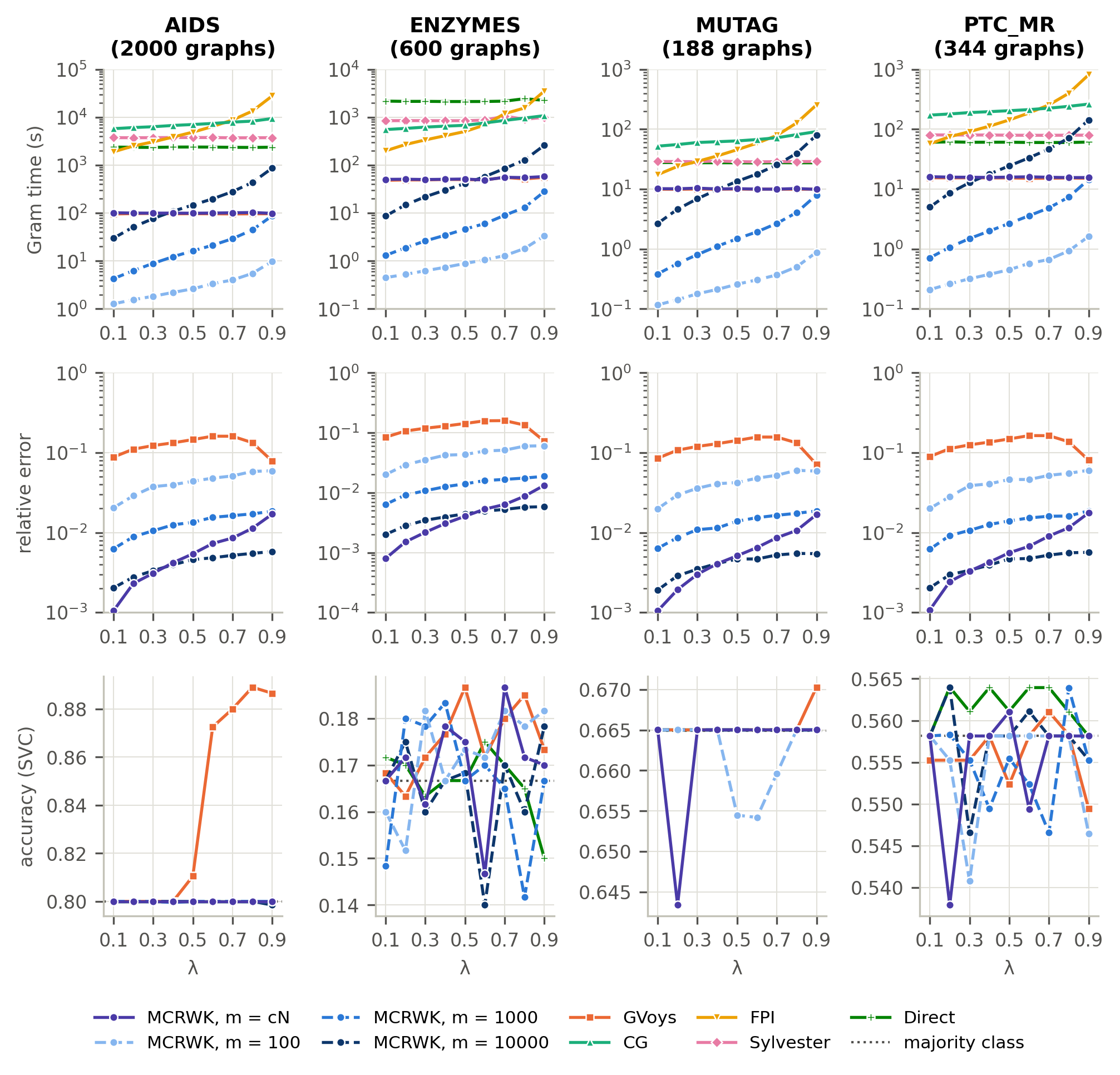}
  \caption{The same as Figure~\ref{fig:lambda_tu_labeled} for unlabeled TU
  datasets.}
  \label{fig:lambda_tu_unlabeled}
\end{figure*}

We repeat the experiments for $\lambda\in\{0.1,0.2,\dots,0.9\}$, while all other
settings are as in Section~\ref{sec:experiments}. The synthetic setting uses pairs of graphs with $N=256$ vertices from the four families of Appendix~\ref{app:setup} and, in the labeled case, three edge labels, each drawn with
probability $1/3$. The reference is CG, since Direct is run only for $N<128$. The TU setting uses the full datasets. Its error is the mean relative
error of the off-diagonal Gram entries against Direct, and its accuracy is the
nested cross-validated SVC accuracy on the normalized Gram matrix. The constant
$c$ of MCRWK with $m=cN$ is calibrated anew for every $\lambda$, so that its
runtime stays matched to GVoys. Figure~\ref{fig:lambda_synthetic} shows the
synthetic results, Figures~\ref{fig:lambda_tu_labeled}
and~\ref{fig:lambda_tu_unlabeled} the TU results.

\paragraph{Runtime.}
Direct and Sylvester do not depend on $\lambda$. The iterative methods slow
down as $\lambda\to1$. FPI needs $O(\log\varepsilon/\log\lambda)$ iterations, and
its Gram time on unlabeled AIDS grows from $1.9\times10^{3}$ to
$2.8\times10^{4}$\,s, for $\lambda\ge0.7$ it is the slowest method. CG
depends on $\lambda$ through the conditioning of $I-\lambda W$ and grows much
less, from $5.8\times10^{3}$ to $9.4\times10^{3}$\,s. Without labels, the cost
of MCRWK at a fixed budget is roughly affine in the expected walk length
$\lambda/(1-\lambda)$. For $m=10^4$ it grows from 30 to 871\,s on AIDS and
from 0.035 to 0.90\,s on a synthetic pair. With $m=10^4$ and $\lambda=0.9$,
MCRWK is therefore slower than Direct on the small datasets MUTAG and PTC\_MR,
while $m=10^3$ remains faster on every dataset. With labels the cost grows much
less, by at most a factor of three, and decreases slightly at $\lambda=0.9$. A
labeled walk stops at the first step whose sampled label does not occur at the
current vertex, so its length is limited by this stopping time rather than by
the sampled length $K$. GVoys and MCRWK with $m=cN$ take a constant time.

\paragraph{Accuracy of the estimate without labels.}
Without labels, the relative error of MCRWK grows only mildly with $\lambda$.
For $m=10^2,10^3,10^4$ it goes from $0.02$, $0.006$ and $0.002$ at $\lambda=0.1$ to $0.06$, $0.019$ and $0.006$ at $\lambda=0.9$. These values
are nearly the same on all four TU datasets and close to the synthetic ones,
so the error is governed by $\lambda$ and $m$ rather than by the graphs. The
error of MCRWK with $m=cN$ grows faster, from $0.001$ to $0.017$, because at a
fixed time budget longer walks leave fewer samples: beyond $\lambda\approx0.4$
its budget falls below $10^4$. The error of GVoys is not monotone in $\lambda$.
It rises from $0.09$ to $0.16$ at $\lambda=0.6$--$0.7$ and falls to $0.08$ at
$\lambda=0.9$. At equal runtime it stays more than an order of magnitude above
MCRWK for every $\lambda\le0.8$, by a factor of 12 to 106, and about 5 times
above it at $\lambda=0.9$.

\paragraph{Accuracy of the estimate with labels.}
With labels, the error of MCRWK grows much faster. For $m=10^4$ it increases
from about $0.003$ at $\lambda=0.1$ to $0.10$ on AIDS and $0.15$ on MUTAG and PTC\_MR at $\lambda=0.9$, and to $0.023$ on the synthetic pairs. With $m=10^2$ it reaches $0.3$--$0.5$ on the TU datasets, which is worse than GVoys. This is
the regime predicted by the variance bound of
Section~\ref{sec:variance_bounds}, which is finite only when $\lambda<q_{\min}$. For
the uniform proposal $q_{\min}=1/L$, where $L$ is the number of edge labels:
$L=3$ for AIDS and the synthetic graphs, $L=4$ for MUTAG and PTC\_MR. Beyond
this threshold the importance weights $\prod_t q(\ell_t)^{-1}$ along a walk become
heavy-tailed, which slows the decay of the error. For large $\lambda$
with labels, a proposal with larger $q_{\min}$ or a larger budget is
needed (Appendix~\ref{ap:q_sampling}).

\paragraph{Classification.}
The SVC accuracy depends little on $\lambda$ on most datasets. Without labels,
the exact kernel does not exceed the fraction of the most frequent class on
AIDS and MUTAG at any $\lambda$, and on ENZYMES and PTC\_MR it stays within about two percentage points of it. MCRWK behaves the same at every budget. With
labels, the exact kernel improves with $\lambda$ on MUTAG (from $0.69$ to
$0.74$) and on AIDS for $\lambda\ge0.8$ (up to $0.85$). MCRWK does not
reproduce these gains: it stays between $0.66$ and $0.71$ on MUTAG and at
$0.80$ on AIDS. This matches its larger error at large $\lambda$ in the labeled
case. GVoys reaches up to $0.90$ on AIDS, above the exact kernel, in both
cases. This gain comes from a size-dependent bias of the GVoys diagonal rather than
from a closer approximation of the kernel (Appendix~\ref{app:tu}). The
biased diagonal of MCRWK gives the same picture.

\subsection{Choice of the boundary vectors}
\label{app:uw}

The kernel depends on the starting and stopping vectors $v$ and $w$. We compare
three choices on the labeled datasets: {normal} (normalized $|N(0,1)|$
draws, used in the other experiments), {uniform} ($v=w=\mathbf 1/N$, used
in Section~\ref{sec:exp-app}) and {degree} ($v=w$ proportional to the
vertex degrees). All runs share the same three seeds, and the setup is
otherwise that of Section~\ref{sec:exp-app}.

\begin{figure*}[t]
  \centering
  \includegraphics[width=\textwidth]{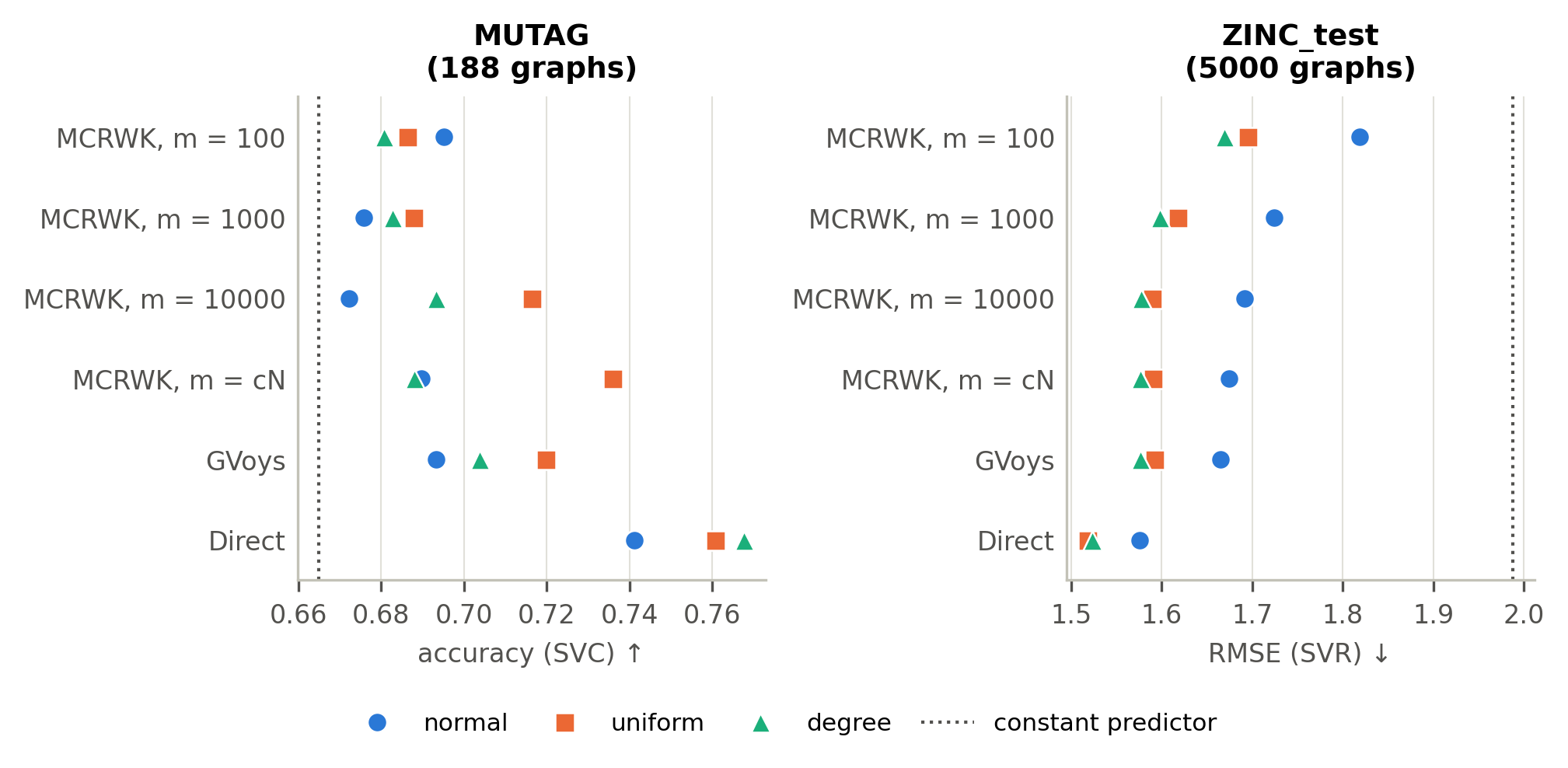}
  \caption{Prediction quality under three choices of the boundary vectors
  $v,w$ on labeled graphs ($\lambda=0.7$), constant predictor is dotted.}
  \label{fig:uw}
\end{figure*}

\begin{table}[t]
  \centering
  \caption{Accuracy (RMSE for ZINC\_test) per choice of $v,w$, labeled graphs,
  $\lambda=0.7$, mean over three seeds.}
  \label{tab:uw}
\fontsize{8}{9.5}\selectfont
  \setlength{\tabcolsep}{3pt}
  \begin{tabular}{lccc@{\hspace{1.2em}}ccc@{\hspace{1.2em}}ccc@{\hspace{1.2em}}ccc}
\toprule
Method & \multicolumn{3}{c}{MUTAG} & \multicolumn{3}{c}{PTC\_MR} & \multicolumn{3}{c}{AIDS} & \multicolumn{3}{c}{ZINC\_test (RMSE)} \\
 & normal & uniform & degree & normal & uniform & degree & normal & uniform & degree & normal & uniform & degree \\
\midrule
MCRWK, $m=10^2$ & 0.695 & 0.686 & 0.681 & 0.539 & 0.550 & 0.549 & 0.800 & 0.800 & 0.800 & 1.819 & 1.695 & 1.669 \\
MCRWK, $m=10^3$ & 0.676 & 0.688 & 0.683 & 0.549 & 0.540 & 0.558 & 0.800 & 0.800 & 0.800 & 1.724 & 1.617 & 1.598 \\
MCRWK, $m=10^4$ & 0.672 & 0.716 & 0.693 & 0.548 & 0.541 & 0.539 & 0.800 & 0.800 & 0.800 & 1.692 & 1.590 & 1.577 \\
MCRWK, $m=cN$ & 0.690 & 0.736 & 0.688 & 0.531 & 0.565 & 0.547 & 0.800 & 0.800 & 0.800 & 1.674 & 1.590 & 1.577 \\
GVoys & 0.693 & 0.720 & 0.704 & 0.556 & 0.550 & 0.559 & 0.902 & 0.878 & 0.900 & 1.665 & 1.592 & 1.577 \\
Exact & 0.741 & 0.761 & 0.768 & 0.550 & 0.547 & 0.550 & 0.801 & 0.800 & 0.800 & 1.576 & 1.518 & 1.523 \\
\midrule
Constant predictor & 0.665 & 0.665 & 0.665 & 0.558 & 0.558 & 0.558 & 0.800 & 0.800 & 0.800 & 1.988 & 1.988 & 1.988 \\
\bottomrule
\end{tabular}

\end{table}

The choice matters for every method. On ZINC\_test, uniform and degree vectors
are clearly better than normal ones: the exact kernel improves from $1.576$ to
$1.518$ and $1.523$, and MCRWK with $m=10^4$ from $1.692$ to $1.590$ and
$1.577$. The gap between MCRWK and the exact kernel also shrinks, from $0.116$
with normal vectors to $0.072$ with uniform and $0.054$ with degree ones.
Random boundary vectors add variation that the walk-based estimators must
resolve, while uniform and degree vectors are smooth and are recovered from
fewer walks.

On MUTAG the same ordering holds for the exact kernel ($0.741$, $0.761$,
$0.768$), but not for the estimators: MCRWK with $m=cN$ reaches $0.736$ with
uniform vectors and only $0.688$--$0.690$ with the other two, and it exceeds
GVoys ($0.720$) only with uniform vectors. On PTC\_MR and AIDS no choice lifts any method above
the constant predictor, and the Gram time differs by less than 5\% between the
three choices.

\subsection{Supervised learning: all datasets and models}
\label{app:tu}

This section completes Section~\ref{sec:exp-app}. It reports the labeled
datasets PTC\_MR and AIDS, the unlabeled case on all datasets, and three further
models: the SVM on the PSD Gram matrix with the biased diagonal, kernel ridge
regression on the Gram matrix, and ridge regression on the explicit features.
The protocol is that of Section~\ref{sec:exp-app}. Ridge is used as a classifier
on one-hot targets, with the regularization parameter selected by the same
nested cross-validation.

\begin{figure*}[t]
  \centering
  \includegraphics[width=\textwidth]{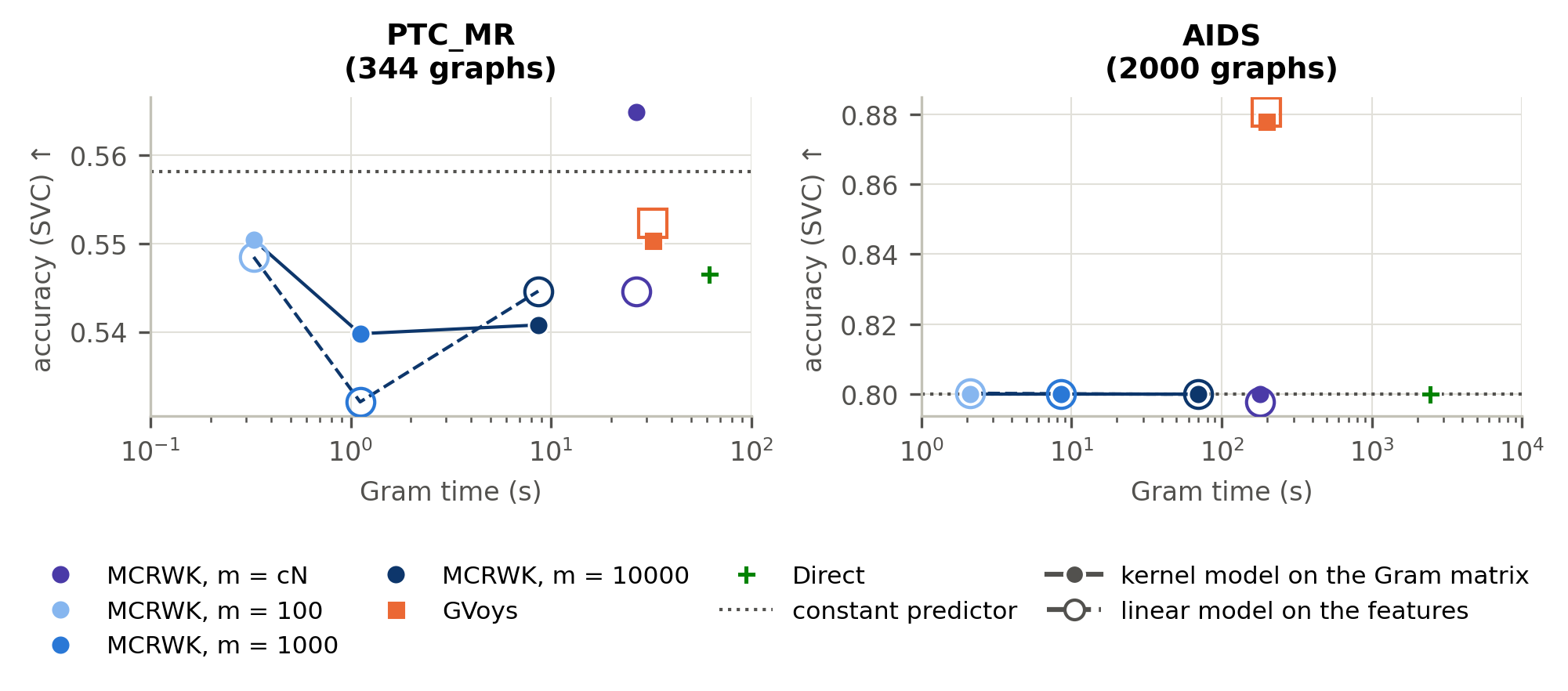}
  \caption{SVC accuracy against the time to build the Gram matrix on the labeled
  datasets PTC\_MR and AIDS ($\lambda=0.7$, uniform $v,w$). Filled markers:
  kernel model on the Gram matrix; hollow markers: linear model on the explicit
  features of GVoys and MCRWK, at the Gram time of the same estimator. Dotted:
  the majority-class predictor.}
  \label{fig:application_appendix}
\end{figure*}

The SVM blocks use the uniform boundary vectors of the main text. The ridge runs
and the whole unlabeled case were computed with normal vectors, and each heading
in Table~\ref{tab:tu_labeled} states which. Appendix~\ref{app:uw} quantifies the
difference, which is a few points of accuracy on MUTAG and about $0.07$ of RMSE
on ZINC\_test. Comparisons between methods are made within a block, never
across two blocks with different vectors. Table~\ref{tab:tu_unlabeled} lists the
unlabeled results, and Figure~\ref{fig:application_appendix} shows PTC\_MR and
AIDS against the Gram time.

\begin{table}[t]
  \centering
  \caption{Labeled graphs, $\lambda=0.7$: accuracy (RMSE for ZINC\_test) of
  every model, mean over three seeds. {Exact} stands for Direct, CG and
  FPI, which give the same kernel. Each heading names the boundary vectors
  used. The biased-diagonal Gram matrix of MCRWK is the inner product of its
  features, so ridge on the features coincides with kernel ridge on that
  matrix.}
  \label{tab:tu_labeled}
  \small
  \begin{tabular}{lcccc}
\toprule
Method & MUTAG & PTC\_MR & AIDS & ZINC\_test (RMSE) \\
\midrule
\multicolumn{5}{l}{\emph{SVM / SVR on the Gram matrix (uniform $v, w$)}} \\
Exact & 0.761 & 0.547 & 0.800 & 1.518 \\
MCRWK, $m=10^2$ & 0.686 & 0.550 & 0.800 & 1.695 \\
MCRWK, $m=10^3$ & 0.688 & 0.540 & 0.800 & 1.617 \\
MCRWK, $m=10^4$ & 0.716 & 0.541 & 0.800 & 1.590 \\
MCRWK, $m=cN$ & 0.736 & 0.565 & 0.800 & 1.590 \\
GVoys & 0.720 & 0.550 & 0.878 & 1.592 \\
\midrule
\multicolumn{5}{l}{\emph{SVM / SVR on the Gram matrix, biased diagonal (uniform $v, w$)}} \\
MCRWK, $m=10^2$ & 0.686 & 0.549 & 0.800 & 1.654 \\
MCRWK, $m=10^3$ & 0.690 & 0.547 & 0.800 & 1.611 \\
MCRWK, $m=10^4$ & 0.718 & 0.543 & 0.800 & 1.588 \\
MCRWK, $m=cN$ & 0.729 & 0.553 & 0.800 & 1.591 \\
GVoys & 0.720 & 0.550 & 0.878 & 1.592 \\
\midrule
\multicolumn{5}{l}{\emph{Linear SVM / SVR on the features (uniform $v, w$)}} \\
MCRWK, $m=10^2$ & 0.695 & 0.548 & 0.800 & 1.649 \\
MCRWK, $m=10^3$ & 0.698 & 0.532 & 0.800 & 1.602 \\
MCRWK, $m=10^4$ & 0.715 & 0.545 & 0.800 & 1.582 \\
MCRWK, $m=cN$ & 0.730 & 0.545 & 0.798 & 1.583 \\
GVoys & 0.721 & 0.552 & 0.881 & 1.582 \\
\midrule
\multicolumn{5}{l}{\emph{Kernel ridge on the Gram matrix (normal $v, w$)}} \\
Exact & 0.738 & 0.536 & 0.814 & 1.556 \\
MCRWK, $m=10^2$ & 0.665 & 0.532 & 0.800 & 1.810 \\
MCRWK, $m=10^3$ & 0.676 & 0.522 & 0.800 & 1.724 \\
MCRWK, $m=10^4$ & 0.695 & 0.547 & 0.799 & 1.695 \\
MCRWK, $m=cN$ & 0.707 & 0.547 & 0.800 & 1.677 \\
GVoys & 0.693 & 0.536 & 0.800 & 1.675 \\
\midrule
\multicolumn{5}{l}{\emph{Ridge on the features (normal $v, w$)}} \\
MCRWK, $m=10^2$ & 0.700 & 0.536 & 0.800 & 1.804 \\
MCRWK, $m=10^3$ & 0.700 & 0.547 & 0.799 & 1.716 \\
MCRWK, $m=10^4$ & 0.692 & 0.554 & 0.800 & 1.684 \\
MCRWK, $m=cN$ & 0.718 & 0.550 & 0.800 & 1.665 \\
GVoys & 0.693 & 0.536 & 0.800 & 1.675 \\
\midrule
Constant predictor & 0.665 & 0.558 & 0.800 & 1.988 \\
\bottomrule
\end{tabular}

\end{table}

\begin{table}[t]
  \centering
  \caption{The same as Table~\ref{tab:tu_labeled} without edge labels, with
  normal boundary vectors}
  \label{tab:tu_unlabeled}
  \small
  \begin{tabular}{lcccccc}
\toprule
Method & MUTAG & PTC\_MR & ENZYMES & AIDS & NCI1 & ZINC\_test (RMSE) \\
\midrule
\multicolumn{7}{l}{\emph{SVM / SVR on the Gram matrix}} \\
Exact & 0.665 & 0.563 & 0.159 & 0.800 & 0.516 & 2.004 \\
MCRWK, $m=10^2$ & 0.658 & 0.557 & 0.171 & 0.800 & 0.490 & 2.005 \\
MCRWK, $m=10^3$ & 0.665 & 0.548 & 0.162 & 0.800 & 0.497 & 2.005 \\
MCRWK, $m=10^4$ & 0.665 & 0.555 & 0.169 & 0.800 & 0.500 & 2.005 \\
MCRWK, $m=cN$ & 0.660 & 0.551 & 0.168 & 0.800 & 0.507 & 2.005 \\
GVoys & 0.667 & 0.546 & 0.187 & 0.873 & 0.576 & 2.005 \\
\midrule
\multicolumn{7}{l}{\emph{SVM / SVR on the Gram matrix, biased diagonal}} \\
MCRWK, $m=10^2$ & 0.665 & 0.558 & 0.186 & 0.800 & 0.495 & 2.005 \\
MCRWK, $m=10^3$ & 0.656 & 0.551 & 0.162 & 0.800 & 0.499 & 2.005 \\
MCRWK, $m=10^4$ & 0.665 & 0.548 & 0.179 & 0.800 & 0.502 & 2.005 \\
MCRWK, $m=cN$ & 0.658 & 0.551 & 0.167 & 0.800 & 0.506 & 2.005 \\
\midrule
\multicolumn{7}{l}{\emph{Linear SVM / SVR on the features}} \\
MCRWK, $m=10^2$ & 0.665 & 0.556 & 0.174 & 0.800 & 0.496 & 1.988 \\
MCRWK, $m=10^3$ & 0.665 & 0.546 & 0.163 & 0.800 & 0.501 & 1.988 \\
MCRWK, $m=10^4$ & 0.665 & 0.554 & 0.181 & 0.800 & 0.502 & 1.988 \\
MCRWK, $m=cN$ & 0.660 & 0.545 & 0.161 & 0.800 & 0.500 & 1.988 \\
GVoys & 0.665 & 0.553 & 0.176 & 0.872 & 0.571 & 1.988 \\
\midrule
\multicolumn{7}{l}{\emph{Kernel ridge on the Gram matrix}} \\
Exact & 0.658 & 0.539 & 0.145 & 0.800 & 0.506 & 1.988 \\
MCRWK, $m=10^2$ & 0.669 & 0.558 & 0.167 & 0.800 & 0.500 & 1.993 \\
MCRWK, $m=10^3$ & 0.665 & 0.548 & 0.155 & 0.799 & 0.508 & 1.990 \\
MCRWK, $m=10^4$ & 0.665 & 0.553 & 0.150 & 0.800 & 0.498 & 1.988 \\
MCRWK, $m=cN$ & 0.665 & 0.545 & 0.173 & 0.800 & 0.502 & 1.988 \\
GVoys & 0.668 & 0.549 & 0.173 & 0.800 & 0.506 & 1.989 \\
\midrule
\multicolumn{7}{l}{\emph{Ridge on the features}} \\
MCRWK, $m=10^2$ & 0.661 & 0.559 & 0.170 & 0.800 & 0.496 & 1.993 \\
MCRWK, $m=10^3$ & 0.661 & 0.552 & 0.168 & 0.800 & 0.501 & 1.989 \\
MCRWK, $m=10^4$ & 0.663 & 0.551 & 0.180 & 0.798 & 0.501 & 1.989 \\
MCRWK, $m=cN$ & 0.663 & 0.548 & 0.155 & 0.799 & 0.506 & 1.988 \\
GVoys & 0.668 & 0.549 & 0.173 & 0.800 & 0.506 & 1.989 \\
\midrule
Constant predictor & 0.665 & 0.558 & 0.167 & 0.800 & 0.500 & 1.988 \\
\bottomrule
\end{tabular}

\end{table}

\paragraph{Labeled PTC\_MR and AIDS.}
On these datasets the exact kernel with SVC does not improve on the constant
predictor. On PTC\_MR every model stays between $0.52$ and $0.57$, around the
majority class ($0.558$): the exact kernel reaches $0.547$ and MCRWK with
$m=cN$ reaches $0.565$, a difference of the order of the spread over the three
seeds. On AIDS the exact kernel with SVC reaches exactly the majority class
($0.800$), and kernel ridge reaches $0.814$. MCRWK gives $0.800$ at every budget
and with every model. The comparison between methods is therefore not
informative on these two datasets, and we do not use them in the main text.

\paragraph{Unlabeled graphs.}
Without edge labels, no model of the exact kernel improves on the constant
predictor on any dataset. On ZINC\_test the RMSE is $1.99$--$2.00$, the
standard deviation of the target. This is a property of the kernel, not of
the estimators. With $W=P_1\otimes P_2$, $v=v_1\otimes v_2$ and $w=w_1\otimes w_2$,
\[
  k(G_1,G_2)=\sum_{k\ge0}\mu_k\,(v^\top W^k w)
  =\sum_{k\ge0}\mu_k\,a_k(G_1)\,a_k(G_2),
  \qquad a_k(G)=v_G^\top P_G^k w_G .
\]
The unlabeled kernel is therefore a weighted inner product of the scalar
sequences $(a_k(G))_{k\ge0}$, one number per walk length. Since $P_G=D^{-1}A$ is
row-stochastic, $P_G^k w_G\to(\pi_G^\top w_G)\mathbf 1$ for a connected,
non-bipartite graph. The sequence is thus determined by a few short walks and
by the boundary vectors. With uniform $w_G$ it is even constant,
$a_k(G)=v_G^\top w_G$, and the normalized Gram matrix has all entries equal to
one, this is why the unlabeled runs use normal vectors. Even then the sequences
carry little structural information. With edge labels the kernel does not
factorize: $W=\sum_\ell P_1^{(\ell)}\otimes P_2^{(\ell)}$, and the features are
indexed by label sequences, of which there are $L^k$ for walks of length $k$.
MCRWK reproduces the unlabeled results: at every budget its accuracy stays
within two percentage points of the exact kernel and of the constant predictor.

\paragraph{Biased diagonal.}
The SVM on the PSD Gram matrix with the biased diagonal performs like the SVM
on the unbiased estimate. Most differences are within the cross-validation
noise, about $\pm0.01$. The exception is the small budget on ZINC\_test, where
the biased diagonal is clearly better: $1.654$ against $1.695$ for $m=10^2$ and
$1.611$ against $1.617$ for $m=10^3$, the gap closing to $0.002$ at $m=10^4$.
This is consistent with its excess $\tau_i\ge0$ on the diagonal, which acts as
an additional ridge and matters most when the estimate is noisiest. The biased
Gram matrix needs no PSD correction and is exactly the Gram matrix of the
features used by the linear models.

\paragraph{Ridge.}
Kernel ridge leads to the same conclusions as the SVM. On ZINC\_test the exact
kernel reaches an RMSE of $1.556$, MCRWK reaches $1.810$, $1.724$ and $1.695$
for $m=10^2,10^3,10^4$ and $1.677$ for $m=cN$, so the ordering in $m$ is the
one of the main text. On labeled MUTAG the exact kernel reaches $0.738$ and
MCRWK with $m=cN$ reaches $0.707$, or $0.718$ on its features. Ridge on the
features and kernel ridge on the biased-diagonal Gram matrix are the same model
in primal and dual form. They agree to three decimals in all our runs, which
numerically confirms that the features of MCRWK reproduce its Gram matrix.

\paragraph{GVoys on AIDS and NCI1.}
With SVC, GVoys reaches $0.878$ on labeled AIDS and $0.873$ on unlabeled AIDS,
well above the exact kernel ($0.800$). It also reaches $0.576$ against $0.516$
on unlabeled NCI1. The linear SVM on its features gives the same values
($0.881$ and $0.571$). The exact kernel stays at the level of the constant
predictor, so this gain does not come from a closer approximation of the random
walk kernel. It comes from graph size. On AIDS the two classes are separated by
the number of vertices alone: every graph with fewer than 15 vertices belongs to
one class, and an SVC on the vertex count reaches $0.997$. The normalized kernel
$K_{ij}/\sqrt{K_{ii}K_{jj}}$ discards this information, but the GVoys diagonal
is inflated by its own sampling variance, more for small graphs (by about 19\%
below 8 vertices and 2\% above 30 on AIDS), and the normalization turns this
bias into a size feature. Normalizing the same GVoys Gram matrix by the exact
diagonal instead brings its accuracy on labeled AIDS ($\lambda=0.7$, normal
$v,w$) from $0.89$ back to $0.80$. The gain is also specific to the hinge loss,
since kernel ridge on the same GVoys Gram matrix and ridge on the same features
give $0.800$ on AIDS and $0.506$ on NCI1. We therefore exclude these datasets
from the comparison in the main text.

\subsection{Effect of edge density}
\label{app:density}

\begin{figure*}[t]
  \centering
  \includegraphics[width=\textwidth]{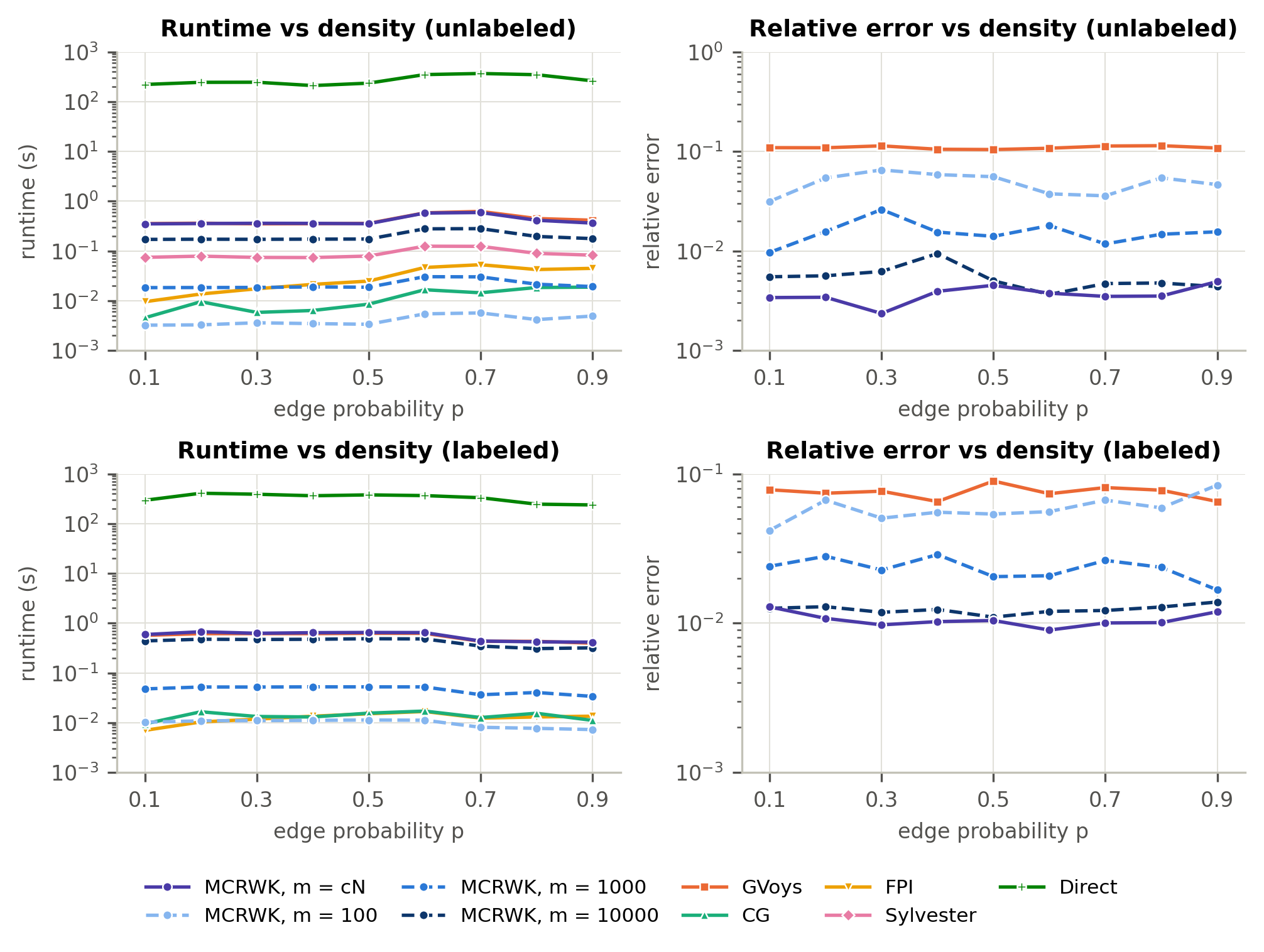}
  \caption{Runtime and relative error of $k(G_1,G_2)$ against the edge
  probability $p$ of Erdos--Renyi graphs with $N=128$ vertices, without and with edge labels, $\lambda=0.7$.}
  \label{fig:density}
\end{figure*}

To separate the effect of the number of edges from that of the number of
vertices, we fix $N=128$ and vary the edge probability of Erdos--Renyi
graphs $G(N,p)$ over $p\in\{0.1,0.2,\dots,0.9\}$, which takes the mean degree
from about 13 to about 114. The remaining setup is that of the scaling
experiment (Section~\ref{sec:scaling-with-graph-size}): $\lambda=0.7$, normal starting and
stopping vectors, three edge labels with probability $1/3$ each in the labeled
case, and only $k(G_1,G_2)$ of each pair is computed. The reference is CG.
Direct, which the other experiments run only for $N<128$, is included here at $N=128$. It
agrees with CG to a relative error below $2\times10^{-10}$. The constant $c$ of
MCRWK with $m=cN$ is calibrated for every $p$. Every point is the mean over 10
pairs, 3 for Direct, whose single pair takes about 4--6 minutes. They vary by up to a factor of 1.6 between neighbouring densities,
uniformly across methods, so we compare methods only at the same $p$.
Figure~\ref{fig:density} shows the results.

\paragraph{Runtime.}
The cost of the walk-based estimators does not depend on the density. Each
step samples one neighbour, whatever the degree. The calibrated constant stays
at $c\approx160$--$166$ without labels and $c\approx103$--$106$ with labels for
every $p$, and the runtimes of GVoys and of MCRWK at a fixed budget change only
with the state of the machine. The iterative exact methods slow down as the
graphs fill in: without labels, CG and FPI become 4 and 4.7 times slower from
$p=0.1$ to $p=0.9$. With labels the growth is smaller, because each
label-specific adjacency matrix holds only a third of the edges. Direct takes
210--410\,s for every $p$, since the LU factors of the $N^2\times N^2$ system
fill in even for sparse graphs. At this size, however, CG, FPI and Sylvester
need only $5\times10^{-3}$ to $0.12$\,s and remain faster than MCRWK with
$m\ge10^3$. The runtime advantage of MCRWK appears with the number of vertices
(Section~\ref{sec:scaling-with-graph-size}), not with the number of edges.

\paragraph{Error.}
The relative error of the estimators does not depend on the density either,
as the variance bounds predict: they involve $\lambda$ and $m$ but not the
graph structure. Without labels, MCRWK has an error of $0.03$--$0.07$,
$0.010$--$0.026$ and $0.004$--$0.009$ for $m=10^2,10^3,10^4$, and
$0.002$--$0.005$ for $m=cN$ ($m\approx2.1\times10^4$). GVoys stays at
$0.10$--$0.11$ at the same runtime. With labels, MCRWK has $0.04$--$0.08$,
$0.017$--$0.029$ and $0.011$--$0.014$, and $0.009$--$0.013$ for $m=cN$, against
$0.065$--$0.09$ for GVoys. The variation across $p$ is within the spread of the
ten repetitions, and none of the curves shows a trend in $p$.

\end{document}